\documentclass[preprint,12pt,authoryear]{elsarticle}

\usepackage{graphicx}

\usepackage{algorithm}
\usepackage{algpseudocode}

\usepackage{booktabs}
\usepackage{multirow}
\usepackage{caption}
\usepackage{rotating}

\usepackage{amssymb}
\usepackage{amsmath}
\usepackage{xcolor}

\definecolor{revcolor}{RGB}{0,0,0} 
\definecolor{delcolor}{RGB}{0,92,184}
\newcommand{\rev}[1]{\textcolor{revcolor}{#1}}

\newenvironment{revblock}{\begingroup\color{revcolor}}{\endgroup}

\usepackage{url}
\usepackage{hyperref}
\hypersetup{
    pdfinfo={
        Title={Industrial3D: A Water-Treatment TLS Point Cloud Dataset and Cross-Paradigm Benchmark for MEP Scene Understanding},
        Author={Chao Yin, Hongzhe Yue, Qing Han, Difeng Hu, Zhenyu Liang, Fangzhou Lin, Bing Sun, Boyu Wang, Mingkai Li, Wei Yao, Jack C.P. Cheng},
    }
}
\usepackage{cleveref}

\usepackage{float}

\usepackage{lineno}
\usepackage{etoolbox}
\AtBeginEnvironment{document}{}
\newcommand{\startmainlinenumbers}{\begin{linenumbers}}
\newcommand{\stopmainlinenumbers}{\end{linenumbers}}

\journal{ISPRS Journal of Photogrammetry and Remote Sensing}

\begin{document}
\hypersetup{pageanchor=false}

\begin{frontmatter}

\title{\rev{Industrial3D: A Water-Treatment TLS Point Cloud Dataset and Cross-Paradigm Benchmark for MEP Scene Understanding}}

\author[label1,label2]{Chao Yin}
\author[label3]{Hongzhe Yue}
\author[label4]{Qing Han}
\author[label5]{Difeng Hu\corref{cor1}}
\author[label1]{Zhenyu Liang}
\author[label1]{Fangzhou Lin}
\author[label1]{Bing Sun}
\author[label7]{Boyu Wang}
\author[label8]{Mingkai Li}
\author[label9,label10]{Wei Yao}
\author[label1]{Jack C.P. Cheng}

\cortext[cor1]{Corresponding author. Email: difenghu@cityu.edu.hk}

\affiliation[label1]{organization={Department of Civil and Environmental Engineering, The Hong Kong University of Science and Technology},
            addressline={Clear Water Bay},
            city={Hong Kong},
            postcode={999077},
            country={China}}
\affiliation[label2]{organization={Guangzhou Institute of Geography, Guangdong Academy of Sciences},
            addressline={100 Xianlie Middle Road},
            city={Guangzhou},
            postcode={510070},
            state={Guangdong},
            country={China}}
\affiliation[label3]{organization={School of Civil Engineering, Southeast University},
            addressline={},
            city={Nanjing},
            postcode={210096},
            country={China}}
\affiliation[label4]{organization={College of Geography and Tourism, Hengyang Normal University},
            addressline={},
            city={Hengyang},
            postcode={421002},
            country={China}}
\affiliation[label5]{organization={Department of Architecture and Civil Engineering, City University of Hong Kong},
            addressline={},
            city={Hong Kong},
            postcode={999077},
            country={China}}
\affiliation[label7]{organization={S.M.A.R.T. Construction Research Group, New York University Abu Dhabi},
            addressline={},
            city={Abu Dhabi},
            postcode={129188},
            country={United Arab Emirates}}
\affiliation[label8]{organization={Department of the Built Environment, National University of Singapore},
            addressline={4 Architecture Drive},
            city={Singapore},
            postcode={117566},
            country={Singapore}}
\affiliation[label9]{organization={Spatial Intelligence and Urban Computing, Institute of Urban Environment, Chinese Academy of Sciences}, city={Xiamen}, country={China}}
\affiliation[label10]{organization={State Key Lab of Ecological Security of Regions and Cities, Institute of Urban Environment, Chinese Academy of Sciences}, city={Xiamen}, country={China}}

\begin{abstract}
\rev{Automated semantic understanding of dense terrestrial laser scanning (TLS) point clouds is a prerequisite for Scan-to-BIM, digital twin maintenance, and as-built verification. Yet for operational industrial mechanical, electrical, and plumbing (MEP) facilities, this challenge remains largely unsolved: water-treatment TLS scans exhibit extreme geometric ambiguity, severe occlusion, and extreme class imbalance that architectural benchmarks such as S3DIS and ScanNet cannot adequately represent. We present Industrial3D, a terrestrial LiDAR dataset with 612.7 million expert-labeled points at 6\,mm resolution from 20 room scenes, 13 dataset areas, and 7 operational water treatment facilities. At 6.6$\times$ the scale of the closest comparable MEP dataset, Industrial3D provides the largest industrial MEP testbed for within-domain scene understanding. We further establish a cross-paradigm benchmark of nine methods across fully supervised, weakly supervised, unsupervised, and foundation-model settings. The best supervised method reaches 55.74\% mIoU, whereas zero-shot Point-SAM reaches 15.79\%, a 39.95 percentage-point gap that quantifies unresolved domain transfer for industrial TLS data. Analysis attributes this gap to a dual crisis: 215:1 statistical rarity and cylindrical geometric ambiguity between tail classes and head-class pipes. The dataset, benchmark code, and pre-trained models will be publicly released at \url{https://github.com/pointcloudyc/Industrial3D}.}
\end{abstract}

\begin{graphicalabstract}
\begin{figure}[H]
\centering
\includegraphics[width=\textwidth]{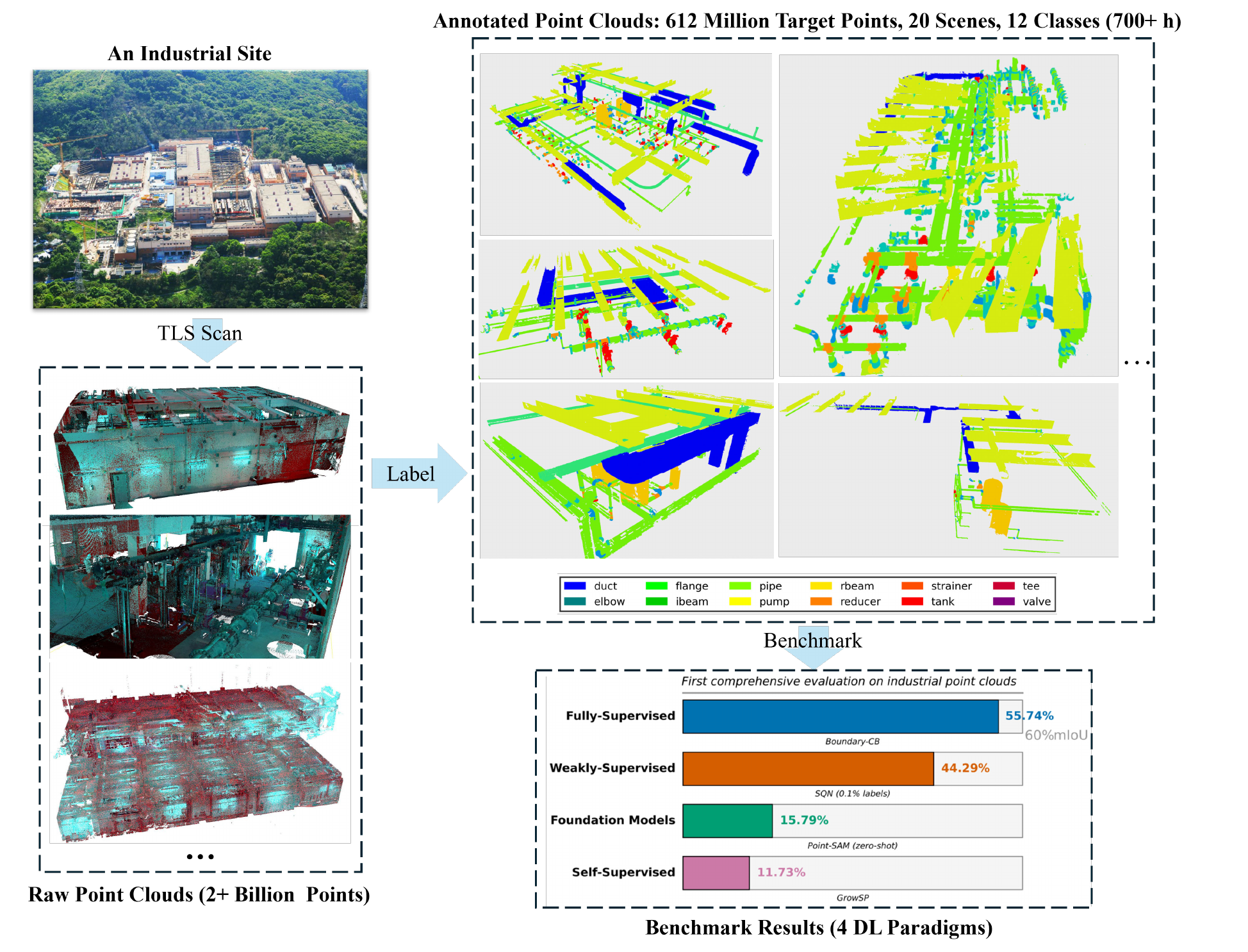}
\end{figure}
\end{graphicalabstract}

\begin{highlights}
    \item \rev{Largest industrial MEP TLS dataset: 612.7M expert-labeled points across 13 areas.}
    \item \rev{Cross-paradigm benchmark evaluating 9 methods across 4 deep learning paradigms.}
    \item \rev{215:1 class imbalance and 39.95 pp zero-shot foundation-model gap quantified.}
    \item \rev{Sparse 0.1\% labels (SQN) match or beat full supervision in the same backbone family.}
\end{highlights}


\end{frontmatter}
\hypersetup{pageanchor=true}

\section{Introduction}
\label{sec:introduction}
Terrestrial laser scanning (TLS) and photogrammetric 3D reconstruction now enable engineers to capture entire industrial facilities as dense, millimetre-accurate point clouds. The resulting data are foundational to Scan-to-BIM pipelines, digital twin construction, and automated as-built verification---tasks that are central to the ongoing digital transformation of the construction industry~\citep{tang2010scan,yin2019,guo2020,wang2022vision}. Realising these applications, however, requires assigning a meaningful semantic label to every point in a scan. This point cloud semantic segmentation step remains a critical bottleneck, and its difficulty is compounded substantially when moving from architectural interiors to operational industrial facilities.

Industrial mechanical, electrical, and plumbing (MEP) environments pose a qualitatively distinct challenge for TLS-based perception. MEP systems can account for up to 40\% of total construction cost~\citep{khanzode2008,lagunas2024nlp}, yet their point cloud representations combine three properties that jointly defeat standard segmentation methods. First, geometrically repetitive components---pipes, elbows, and reducers---are locally indistinguishable from one another; most of tail-class points share cylindrical primitive geometry with head-class pipes, creating systematic ambiguity that frequency-based re-weighting cannot resolve. Second, dense equipment arrangements produce severe mutual occlusion, leaving many components only partially captured and sampled. Third, the class frequency distribution is extreme: the head-to-tail imbalance in operating water treatment plants reaches 215:1, a ratio 3.5$\times$ more severe than indoor architectural benchmarks. Together, these unique properties constitute what this paper terms a \textit{dual crisis} of statistical rarity and geometric ambiguity.

Progress on this problem is constrained by the absence of appropriate benchmarks. The most widely used 3D scene understanding datasets---S3DIS~\citep{armeni2016} and ScanNet~\citep{dai2017}---target offices, corridors, and domestic interiors captured with RGB-D sensors, creating a double domain gap (scene type and sensor modality) when their trained models are transferred to TLS acquisitions of industrial plants. MEP-focused datasets exist, but \citet{jing2024mep} is limited to relatively smaller scale (92.1M points) from commercial and residential buildings, and no prior dataset captures the geometric complexity, imbalance severity, \rev{and subsystem variation of operating water-treatment facilities. Furthermore, existing datasets lack the scale and scene diversity required to fairly evaluate modern learning paradigms such as sparse supervision at the 0.1\% label level, held-out area generalisation}, and zero-shot foundation model transfer. Without a benchmark grounded in real industrial acquisition conditions, the community cannot measure how large the industrial domain gap truly is, nor systematically evaluate strategies for closing it.

This paper addresses these limitations by introducing the Industrial3D dataset and a cross-paradigm benchmark for industrial point cloud segmentation. \rev{Industrial3D contains 612.7 million expertly labelled points from 20 room scenes grouped into 13 dataset areas across 7 operational} water treatment facilities in Hong Kong, captured at 6\,mm TLS resolution and labelled into 12 specialised MEP and structural classes. \rev{The rooms span chiller plants, service galleries, ozone and sodium-chloride generation rooms, biological filtration areas, sludge press houses, pumping rooms, and washwater recovery zones, so dataset value derives from scene and subsystem diversity together with held-out area generalisation, rather than from total point count alone}. At 6.6$\times$ the scale of the closest comparable MEP dataset, it provides a demanding \rev{water-treatment MEP testbed}. The benchmark evaluates nine representative methods under a unified benchmark protocol across four learning paradigms: fully supervised, weakly supervised, unsupervised, and zero‑shot foundation models. Results reveal the full extent of the dual crisis: the best supervised method achieves 55.74\% mIoU, yet zero-shot Point-SAM reaches only 15.79\%---a 39.95 percentage-point gap that quantifies the unresolved domain-transfer challenge. The benchmark further demonstrates that sparse supervision can match or exceed full supervision within the same architectural family, and that current foundation models lack the representations needed for industrial MEP semantics.

The contributions of this work are as follows:

\begin{enumerate}
    \item We present Industrial3D, the largest terrestrial LiDAR dataset \rev{prepared for public release for water-treatment} MEP scene understanding: 612.7 million labelled points at 6\,mm TLS resolution across 12 specialised classes, \rev{20 room scenes, and 13 dataset areas from 7 operational} water treatment facilities, 6.6$\times$ larger than the closest comparable MEP dataset~\citep{jing2024mep}. \rev{Its room and subsystem diversity} enables studies that were previously infeasible, including 0.1\% sparse-supervision experiments, cross-area generalisation across multiple facilities, and \rev{downstream MEP modelling protocols} that demand the separation of functionally distinct but geometrically similar components.

    \item We establish the first industrial cross-paradigm benchmark, evaluating nine representative methods across fully supervised, weakly supervised, unsupervised, and foundation model settings under a unified PyTorch codebase. Methods originally implemented in TensorFlow, including RandLA-Net~\citep{hu2020randlanet} and SQN~\citep{hu2022sqn}, were re-implemented and validated to ensure reproducible comparison.

    \item We provide a systematic analysis of three persistent challenges in industrial point cloud segmentation---extreme class imbalance, sparse supervision, and foundation model domain adaptation---connecting benchmark results to their underlying causes and clarifying which obstacles are architectural, data-driven, or transfer-related.
\end{enumerate}

The remainder of this paper is organised as follows. \Cref{sec:related_work} reviews relevant datasets and learning paradigms. \Cref{sec:dataset} describes Industrial3D's acquisition, annotation, and statistical characteristics. \Cref{sec:benchmark} presents the benchmark design and quantitative results. \Cref{sec:challenges} analyses the three industrial-specific challenges exposed by those results. \Cref{sec:discussion} discusses implications and limitations, and~\Cref{sec:conclusion} concludes the paper.

\section{Related Work}
\label{sec:related_work}

Point cloud semantic segmentation has advanced through better datasets, stronger architectures, and a broader set of learning paradigms. This section reviews prior work along those three dimensions to position Industrial3D. Across all three strands, the primary gap is not a shortage of segmentation methods but the absence of realistic, large-scale benchmarks for industrial point clouds that enable fair cross-paradigm comparison.

\subsection{3D Scene Understanding Datasets}
Operating industrial facilities represent a fundamentally distinct scenario from the scenes targeted by dominant benchmarks: higher component density, more severe mutual occlusion, and no stable reference geometry such as walls or floors. S3DIS~\citep{armeni2016} provides 273M points in 271 rooms with 13 classes centred on walls, floors, and furniture, while ScanNet~\citep{dai2017} expands indoor coverage to roughly 1,500 scans and 20 classes but retains the same architectural emphasis. Both rely on RGB-D sensors rather than TLS, introducing an additional sensor-domain gap. Outdoor datasets such as Semantic3D~\citep{hackel2017} and SemanticKITTI~\citep{behley2019semantickitti} are large and influential yet target urban or driving scenes rather than built industrial environments.

Construction-specific datasets remain limited in scope. CLOI~\citep{agapaki2019cloi} contains 12,125 industrial objects across 140M TLS points but focuses on object-level shape classification (cylinders, elbows, I-beams) rather than scene-level MEP semantic segmentation. PSNet5~\citep{yin2021}, the first industrial MEP dataset, introduced 80M points across 5 classes from 3 water treatment facilities. More recent datasets include Building-PCC~\citep{gao2024buildingpcc} for point cloud completion and Rohbau3D~\citep{rauch2025rohbau3d} for construction shell phases, but neither addresses operational MEP systems.

Most relevant to Industrial3D is the dataset by \citet{jing2024mep}, which contributes 92.1M labelled points across 9 MEP classes from 4 commercial and residential buildings. Commercial and residential buildings, however, lack the dense equipment arrangements, extreme occlusion patterns, and operational complexity that characterise water treatment plants, and this dataset is 6.6$\times$ smaller than Industrial3D.

Existing datasets share three limitations for industrial MEP understanding: (1) \textit{domain mismatch}---specialised classes such as valves, flanges, and pumps are absent or severely underrepresented; (2) \textit{insufficient scale}---point counts are too low to support modern evaluation settings such as 0.1\% sparse supervision or foundation model transfer; and (3) \textit{limited scene diversity}---coverage is restricted to \rev{narrow building or facility contexts, undermining within-domain generalisation assessment}.

\Cref{tab:related_work_comparison} summarises this comparison. Industrial3D addresses these gaps as the largest industrial MEP dataset \rev{prepared for public release, combining scale (612.7M points), fine-grained semantics (12 classes), authentic TLS acquisition, subsystem diversity across 20 room scenes and 7 operational facilities}, and the first cross-paradigm evaluation on industrial point clouds.

\begin{table}[htbp]
\centering
\caption{Comparison of Industrial3D with existing point cloud datasets.
Acq.\ = acquisition sensor; M = million, B = billion; * = not publicly available.
Industrial3D is the largest TLS-based industrial dataset with semantic labels.}
\label{tab:related_work_comparison}
\resizebox{\linewidth}{!}{%
\begin{tabular}{@{}llccccc@{}}
\toprule
\textbf{Dataset} & \textbf{Domain} & \textbf{Acq.} & \textbf{Year} & \textbf{Pts} & \textbf{Cls} & \textbf{Imb} \\
\midrule
S3DIS~\citep{armeni2016} & Indoor & RGB-D & 2017 & 273M & 13 & 62:1 \\
ScanNet~\citep{dai2017} & Indoor & RGB-D & 2017 & 242M & 20 & $\sim$25:1 \\
Semantic3D~\citep{hackel2017} & Outdoor & TLS & 2017 & 4B & 8 & 75:1 \\
SemanticKITTI~\citep{behley2019semantickitti} & Driving & MLS & 2019 & 4.5B & 28 & $\sim$150:1 \\
\midrule
CLOI*~\citep{agapaki2019cloi} & Industrial & TLS & 2019 & 140M & 10 & -- \\
PSNet5~\citep{yin2021} & Industrial & TLS & 2021 & 80M & 5 & $\sim$50:1 \\
\citet{jing2024mep} & MEP (Common Bldg.) & TLS & 2024 & 92.1M & 9 & -- \\
Rohbau3D~\citep{rauch2025rohbau3d} & Construction Shell & TLS & 2025 & 504 scans & -- & -- \\
\textbf{\rev{Industrial3D (ours)}} & \textbf{\rev{MEP (Industrial)}} & \textbf{\rev{TLS}} & \textbf{\rev{2026}} & \textbf{\rev{612M}} & \textbf{\rev{12}} & \textbf{\rev{215:1}} \\
\bottomrule
\end{tabular}%
}
\end{table}

\subsection{Scan-to-BIM and As-Built Verification}
\label{sec:scan_bim_related}
\begin{revblock}
Scan-to-BIM research converts reality-capture data into geometric and semantic representations that can be compared with design intent, used for facility management, or integrated into digital-twin workflows~\citep{tang2010scan,ma2018reconstruction}. Prior work has progressed from rule-based primitive extraction to learning-assisted MEP reconstruction and BIM-to-scan adaptation, including early deep-learning Scan-to-BIM frameworks for complex MEP scenes~\citep{yin2019,wang2022vision,hu2024automated}. Recent GeoBIM and as-designed BIM integration studies further show that semantic LiDAR point clouds can support as-built-versus-design checking when coordinate alignment, object correspondence, and component-level semantics are available~\citep{shao2024urban}.

These downstream workflows require more than class labels. They need reliable separation of adjacent component instances, primitive parameters for pipes and ducts, and an as-designed BIM reference with a documented alignment protocol. The present Industrial3D revision provides semantic point labels as the benchmark target, while as-designed BIM/IFC references are internally available for at least three scanned room scenes. These BIM files are third-party facility and design-contractor assets and are not included in the public release planned for this revision. \rev{We therefore position Scan-vs-BIM as authorised future work, pending owner-approved BIM access, rather than claiming a complete reconstruction benchmark or public Scan-vs-BIM benchmark (see Discussion in \Cref{sec:discussion}).} This positioning keeps the current contribution measurable while preparing a path for downstream verification when access, anonymisation, room-model mapping, and alignment conditions are approved.
\end{revblock}

\subsection{Fully-Supervised Methods}
\label{sec:fully_supervised_methods}
Deep learning has driven significant progress in point cloud semantic segmentation. As comprehensively surveyed by \citet{guo2020} and, for construction applications specifically, by \citet{yue2024deep}, point-based, voxel-based, and projection-based approaches have evolved rapidly, establishing strong theoretical foundations for 3D understanding~\citep{HUANG202062}.

Point-based methods operate directly on unordered point sets. PointNet~\citep{qi2017pointnet} pioneered permutation-invariant architectures using shared MLPs and max pooling for global feature aggregation. PointNet++~\citep{qi2017pointnetpp} extended this with hierarchical feature learning through progressive neighbourhood grouping. Recent advances include kernel-based convolutions (KPConv~\citep{thomas2019}), efficient random sampling strategies (RandLA-Net~\citep{hu2020randlanet}), and serialised attention mechanisms (Point Transformer V3~\citep{wu2024point}). Voxel-based methods, typified by SparseConvNet~\citep{graham2017}, convert point clouds to sparse 3D grids for efficient convolution but introduce discretisation artefacts. Projection-based methods process multi-view 2D renderings using mature CNN architectures at the cost of inherent 3D geometric information.

Construction-specific applications have explored these architectures for infrastructure analysis. \citet{pierdicca2020} applied PointNet to heritage architecture, \citet{agapaki2020} developed CLOI-NET for industrial component detection, and \citet{yin2022} introduced SE-PseudoGrid for piping system classification from LiDAR data. Our prior works, ResPointNet++~\citep{yin2021}, integrated residual learning into PointNet++ for real-world industrial MEP point clouds spanning 5 classes and \citet{jing2026} proposed Building-MLLM, a multimodal large language model for semantic understanding of indoor building components.

Although these studies demonstrate the applicability of supervised methods to construction and industrial domains, they address architectural elements, synthetic environments, or limited class taxonomies. \rev{The dual crisis of statistical rarity and geometric ambiguity that characterises operating industrial facilities remains insufficiently measured under consistent benchmark conditions. Industrial3D provides such a measurement setting}. Fully supervised methods, however, require complete annotation, motivating the label-efficient alternatives reviewed next.

\subsection{Weakly-Supervised Methods}
\label{sec:weakly_supervised_methods}
The annotation cost for large-scale industrial datasets is prohibitive, making label-efficient learning an important direction. \rev{Following \citet{zhou2018brief} and the 3D point-cloud label-efficiency survey by \citet{xiao2024survey}}, weakly supervised learning (WSL) methods are categorised by supervision type: \textit{incomplete} (a subset of data labelled), \textit{inexact} (coarse labels such as scribbles), and \textit{inaccurate} (noisy labels). For point clouds, incomplete supervision---sparse point labelling within scenes---is most prevalent. Existing approaches include 2D label transfer~\citep{wang2019towards}, superpoint-based annotation reduction~\citep{landrieu2018large}, and contrastive pretraining with sparse fine-tuning~\citep{xie2020pointcontrast}. Sparse labels, active learning, and pseudo-labelling have also been shown to substantially reduce supervision demand in geospatial point cloud segmentation~\citep{wang2022als,wang2023one}. Under the incomplete supervision paradigm, consistency learning~\citep{xu2020weakly,zhang2021perturbed} and pseudo-label self-training~\citep{liu2021oneclick,yin2023} are common strategies for propagating sparse annotations.

Semantic Query Network (SQN)~\citep{hu2022sqn} is representative of the state of the art, exploiting local semantic homogeneity through a direct query mechanism that propagates sparse supervision across wider spatial regions than conventional U-Net hierarchies. Cross-Point Contrastive Learning (CPCM)~\citep{liu2023cpcm} further improves label-efficient performance via contextual contrastive learning. \rev{Their behaviour on specialised domains with extreme class imbalance and geometric ambiguity---such as industrial MEP---has not been systematically evaluated under a shared industrial protocol}. Label-efficient methods reduce but do not eliminate annotation requirements, therefore fully annotation-free approaches are desired.

\subsection{Unsupervised Methods}
\label{sec:unsupervised_methods}
Unsupervised representation learning exploits abundant unlabelled data without manual annotation, making it attractive for industrial applications where expert labelling is costly. The dominant paradigm is contrastive learning, in which representations are trained by pulling features of similar point cloud views together and pushing dissimilar views apart. PointContrast~\citep{xie2020pointcontrast} and OcCo~\citep{wang2021occo} demonstrated that features learned via such pretext tasks transfer effectively to downstream segmentation. PointDC~\citep{chen2023pointdc} strengthens this through cross-modal distillation and super-voxel clustering, while GrowSP~\citep{zhang2023growsp} forgoes contrastive pairs entirely, learning semantics by clustering low-level geometric features and progressively growing superpoints. More recent methods such as Concerto~\citep{zhang2025concerto} and Sonata~\citep{wu2025sonata} introduce multi-modal contrastive learning and diagnose the geometric shortcut problem---the tendency of geometry-only models to collapse to low-level structural features rather than semantic representations.

The principal limitation of current unsupervised methods is precisely the geometric shortcut problem: when tail-class points are geometrically indistinct from head-class pipes, \rev{contrastive or clustering objectives on geometry alone may not separate them}. In this study, Industrial3D provides a downstream benchmark for assessing how well unsupervised representations handle the fine-grained and highly specific geometry of industrial MEP components. \rev{Whether such methods transfer meaningfully to industrial scenes is quantified in \Cref{sec:benchmark}.}

\subsection{Foundation Models}
\label{sec:foundation_models}
Foundation models pre-trained on large 3D corpora raise the possibility of zero-shot or few-shot transfer, which is attractive for industrial applications where annotation is expensive. Point-SAM~\citep{zhou2025pointsam} adapts the Segment Anything paradigm to point clouds through prompt-based segmentation, PointCLIP~\citep{zhang2022pointclip} transfers vision-language supervision to 3D object understanding, and Reason3D~\citep{huang2025reason3d} continues the trend towards larger pre-training corpora and broader transfer.

\begin{revblock}
A critical open question is whether these models transfer to industrial semantics rather than merely to new architectural benchmarks. The pretrained distributions of current 3D foundation models are dominated by indoor scenes (e.g., S3DIS and ScanNet) with planar geometry (walls, floors, furniture), which differs substantially from the cylindrical-primitive geometry and MEP-specific semantics of industrial facilities. In this study, Industrial3D measures this domain gap directly, as reported in \Cref{sec:benchmark}.
\end{revblock}

\begin{table}[htbp]
\centering
\caption{\rev{Summary of the method families reviewed in \Cref{sec:fully_supervised_methods,sec:weakly_supervised_methods,sec:unsupervised_methods,sec:foundation_models} and their relevance to Industrial3D.}}
\label{tab:related_method_summary}
{\color{revcolor}
\small
\begin{tabular}{@{}p{2.4cm}p{2.7cm}p{3.0cm}p{3.5cm}@{}}
\toprule
\textbf{Family} & \textbf{Typical supervision} & \textbf{Strength} & \textbf{Open issue for industrial MEP scenes} \\
\midrule
Fully supervised segmentation & Dense point labels & Strong per-point accuracy on labelled domains & Requires expensive expert labels and still struggles with rare, geometrically ambiguous fittings \\
Weakly supervised segmentation & Sparse points, scribbles, or incomplete labels & Reduces annotation burden through label propagation & Sparse labels may not cover rare classes or sharp MEP boundaries \\
Unsupervised representation learning & No semantic labels & Uses abundant unlabelled scans & Geometry-only objectives can collapse to pipes and beams when functional semantics share primitives \\
Foundation models & Large-scale pre-training plus prompts or few-shot adaptation & Enables zero-shot or few-shot transfer & Pre-training is dominated by architectural scenes, leaving industrial MEP semantics out of distribution \\
\bottomrule
\end{tabular}
}
\end{table}

\section{Industrial3D Dataset}
\label{sec:dataset}

We introduce Industrial3D, a terrestrial LiDAR dataset designed for semantic segmentation in industrial MEP scenes. Collected from operational water treatment facilities in Hong Kong, it contains 612.7 million expert-labelled points across 12 semantic classes. This section describes data collection and preprocessing (\Cref{sec:collection}), annotation methodology (\Cref{sec:annotation}), dataset statistics and characteristics (\Cref{sec:statistics}), and comparison with existing datasets (\Cref{sec:comparison}).

\subsection{Data Collection and Preprocessing}
\label{sec:collection}
We captured Industrial3D in operational water treatment facilities in Hong Kong. This domain was selected because it is both underexplored and methodologically challenging: water treatment plants contain dense piping networks, many embedded components, and severe mutual occlusion, all of which complicate semantic understanding. Water treatment is also a ubiquitous infrastructure type worldwide, so the resulting benchmark is domain-specific without being site-specific.

\begin{revblock}
The dataset spans 20 room scenes grouped into 13 distinct dataset areas across 7 operational water treatment facilities, following the S3DIS area-based evaluation protocol. This hierarchy separates facility identity, dataset-area split, and room-level scenes: some dataset areas contain multiple rooms, for example Area~6 includes the 93m PSU and 99.2 VSPA-1 rooms. These areas cover chiller plants, service galleries, ozone and sodium-chloride generation rooms, biological filtration areas, pumping rooms, sludge press houses, and washwater recovery zones, collectively representing a range of water-treatment MEP environments encountered in practice.
\end{revblock}

\begin{figure}[H]
    \centering
    \includegraphics[width=0.98\textwidth]{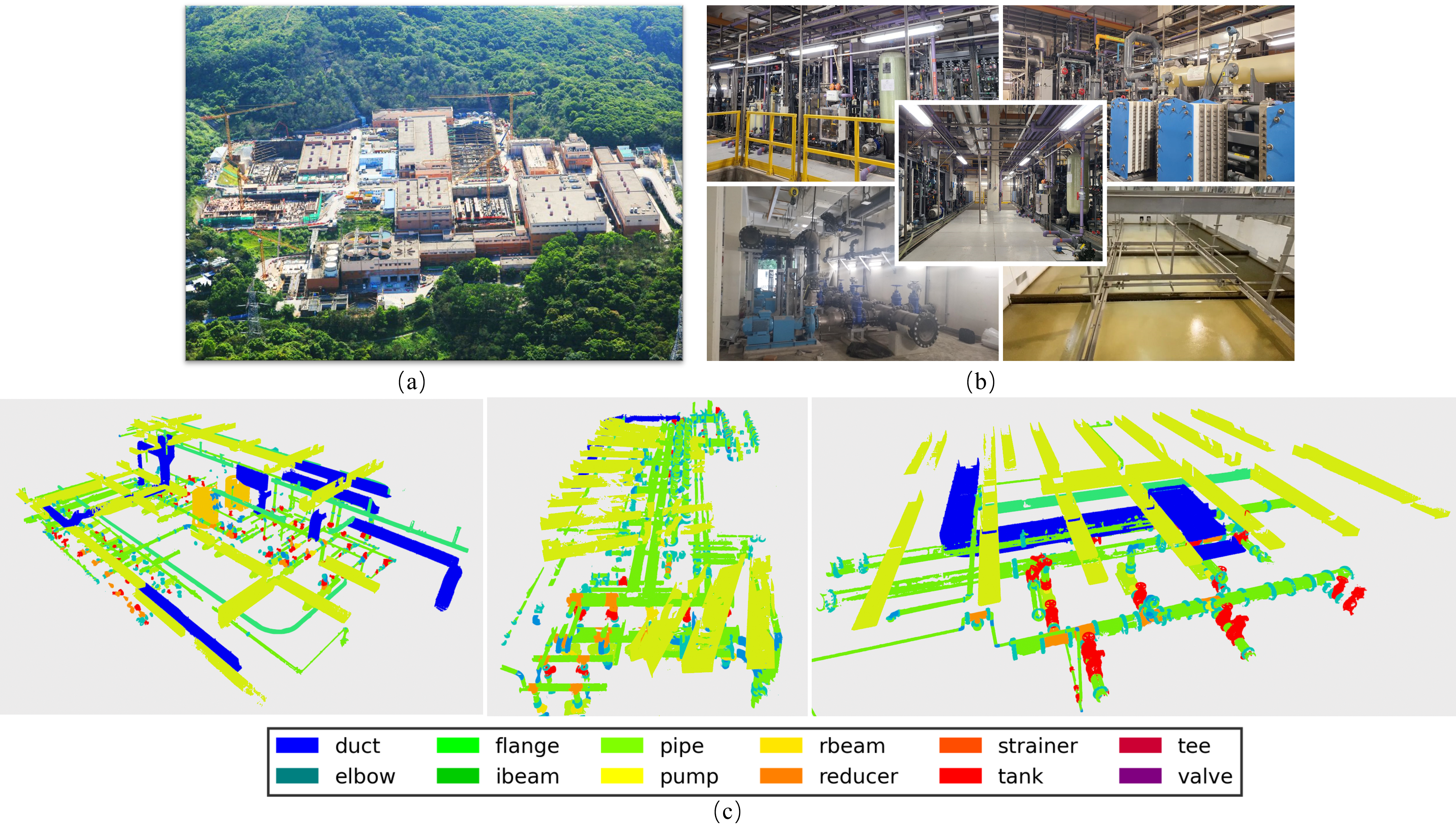}
    \caption{Overview of the Industrial3D dataset from representative water treatment facilities. (a) UAV image of the facility. (b) On-site photographs from five representative rooms (Areas 1, 2, 9, 12, and 13). (c) Manually annotated point clouds for three representative scenes showing 12-class MEP labels; structural elements and non-MEP clutter are omitted for clarity.}
    \label{fig:overview}
\end{figure}

\rev{Several selected facilities employ BIM in design and operation, making them natural internal testbeds for Scan-to-BIM research under owner-approved access}. Unlike staged construction sites, these operational plants exhibit the geometric complexity produced by years of expansion and retrofitting. A representative facility illustrates this pattern: a capacity-doubling expansion was executed within the existing footprint, forcing dense multi-loop piping layouts that directly produce the geometric ambiguity and component density that define Industrial3D's segmentation difficulty. \Cref{tab:room_stats} summarises the physical dimensions of each scene, and~\Cref{fig:overview} provides a visual overview.

\begin{table}[htbp]
\centering
\caption{Scene-level physical dimensions of representative scene types in Industrial3D. Length, width, and floor area (m, m, m\textsuperscript{2}) for each scene, totalling 11,237\,m\textsuperscript{2} \rev{across seven measured representative room/scene types}. Area abbreviations: OG=Ozone Generation, OSCG=Ozone/Sodium Chloride Generation, PBF=Primary Biological Filter, PSU=Pumping Station Unit, SPH=Sludge Press House, VPA=Variable Pump Area, WRT=Washwater Recovery Tank.}
\label{tab:room_stats}
\begin{tabular}{@{}lrrr@{}}
\toprule
\textbf{Scene Name} & \textbf{Length (m)} & \textbf{Width (m)} & \textbf{Area (m²)} \\ \midrule
Chiller & 45.1 & 19.2 & 865.92 \\
OG & 86.2 & 42.5 & 3,663.50 \\
OSCG & 15.3 & 38.2 & 584.46 \\
PBF-1 & 58.0 & 22.0 & 1,276.00 \\
PBF-2 & 68.4 & 38.2 & 2,612.88 \\
SPH & 36.1 & 26.9 & 971.09 \\
WRT2 & 53.3 & 23.7 & 1,263.21 \\ \midrule
\textbf{Total} & - & - & \textbf{11,237.06} \\ \bottomrule
\end{tabular}
\end{table}

\begin{figure}[H]
    \centering
    \includegraphics[width=0.6\textwidth]{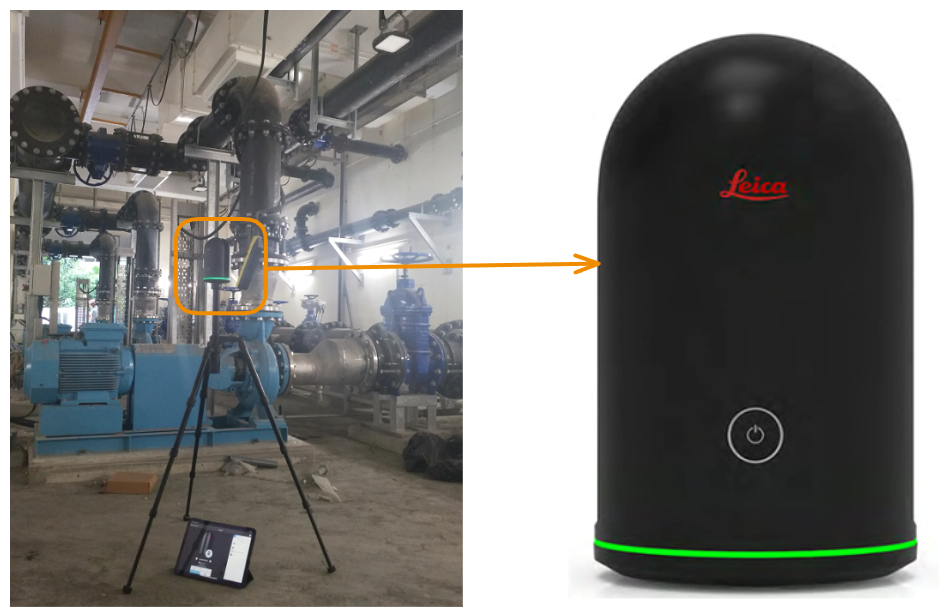}
    \caption{Data acquisition equipment for Industrial3D. The Leica BLK360 terrestrial laser scanner is shown in detail (inset) and deployed in an industrial scene (WRT2), where an iPad Pro controller facilitates real-time preview and registration.}
    \label{fig:industrial3d_scanning}
\end{figure}

We acquired data using a Leica BLK360 terrestrial laser scanner (TLS), as shown in~\Cref{fig:industrial3d_scanning}. \rev{TLS was preferred over mobile laser scanning, photogrammetry, and structured-light (Kinect-like) sensors because its static, high-density acquisition provides the geometric fidelity and resolution needed to resolve small, densely packed, and partially occluded MEP fittings.} The instrument operates at 360,000 points per second with 6\,mm point spacing at a 10\,m range \rev{and a ranging accuracy of 4\,mm at 10\,m (7\,mm at 20\,m)}, providing sufficient resolution to resolve small MEP components such as valves and flanges (typically 50--100\,mm in diameter). \rev{The scanner records seven attributes per point at acquisition---3D coordinates (XYZ), colour (RGB), and return intensity; the released benchmark files provide the six XYZRGB attributes as standard input modalities.} \rev{Scan stations were planned from the available walkways, corners, and equipment clearances in each room to maximise line-of-sight coverage of pipe racks and embedded devices while preserving safe access in operating facilities. Operators used the scanner preview and room-level walk-throughs to add stations for occluded pipe loops, valve clusters, and pump assemblies, with each room typically requiring several hours of multi-station acquisition to achieve adequate coverage. The exact number of stations and registration residuals were not consistently preserved in the historical project metadata, so we do not report unverifiable station-count or RMSE values.}

We processed raw scan data using Autodesk ReCap in three steps: (1) \textbf{registration}, where we aligned multi-station scans into a single coherent point cloud via cloud-to-cloud iterative closest point alignment; \rev{(2) \textbf{quality review}, where registered rooms were visually inspected for gross misalignment, missing high-value MEP zones, and duplicate or isolated scan fragments; and (3) \textbf{export}, where we converted the registered data from the proprietary scanner format to open TXT and PLY formats compatible with annotation tools and deep learning frameworks. The released semantic benchmark contains the labelled MEP and structural classes; unclassified background and unidentifiable objects are retained in the source processing records but excluded from metric computation.}

\subsection{Data Annotation}
\label{sec:annotation}
Five domain experts created annotations for the dataset over approximately 754 person-hours (\Cref{tab:annotation_hours}). This substantial investment reflects both the scale of the dataset (612.7M points across 13 areas) and the geometric complexity of industrial MEP scenes, where many components are partially occluded and require domain knowledge to identify correctly.

\begin{figure}[H]
    \centering
    \includegraphics[width=0.85\textwidth]{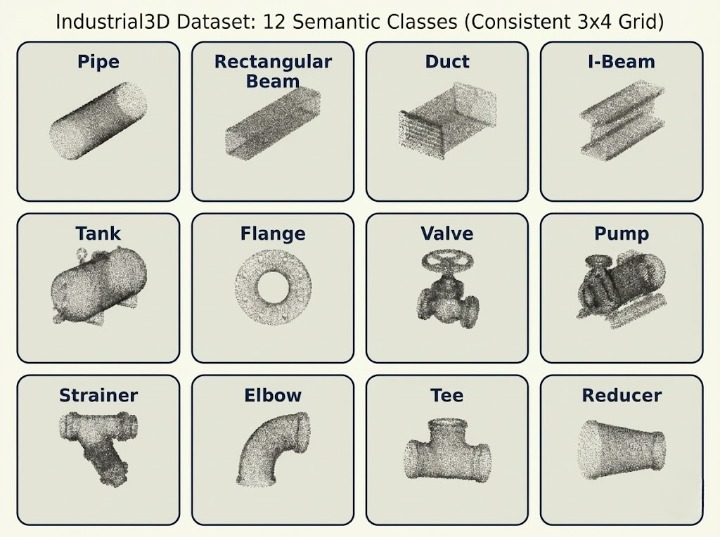}
    \caption{Representative instances of each semantic class in Industrial3D, rendered as synthetic point clouds from BIM models for clarity. Actual TLS scans exhibit significant occlusion, noise, and incomplete coverage. Pipe-related classes (pipe, elbow, tee, reducer) share geometric similarity that creates segmentation challenges.}
    \label{fig:industrial3d_examples}
\end{figure}

Drawing on established MEP datasets and industry standards~\citep{agapaki2019cloi,yin2021,yeo2020mep}, we defined a taxonomy of 12 semantic classes covering both MEP and structural elements, as detailed in~\Cref{tab:industrial3d_classes}. We assigned points corresponding to floors, ceilings, walls and unidentifiable objects to an \texttt{unclassified} category and excluded them from evaluation. \Cref{fig:industrial3d_examples} illustrates representative instances of each semantic category. Note that these visualizations are synthetic point clouds sampled from BIM models, not actual LiDAR scans; real data exhibit substantial occlusion, noise, and incomplete coverage due to terrestrial scanning geometry.

\begin{table}[htbp]
\centering
\caption{The 12 semantic classes in Industrial3D, organised by frequency tier (Head: 77\%, Common: 20\%, Tail: 3\% of points).}
\label{tab:industrial3d_classes}
\small
\begin{tabular}{@{}llp{7.5cm}@{}}
\toprule
\textbf{Abbrev.} & \textbf{Class Name} & \textbf{Description} \\
\midrule
\multicolumn{3}{l}{\textit{Head classes (77\% of points)}} \\
\quad Pipe & Pipe & Cylindrical conduits for fluid transport \\
\quad Rbm & Rectangular beam & Rectangular structural support members \\
\quad Dct & Duct & Rectangular air-handling conduits \\
\midrule
\multicolumn{3}{l}{\textit{Common classes (20\% of points)}} \\
\quad Ibm & I-beam & I-shaped structural steel beams \\
\quad Tank & Tank & Large cylindrical or rectangular storage vessels \\
\midrule
\multicolumn{3}{l}{\textit{Tail classes ($<$3\% each)}} \\
\quad Flg & Flange & Circular connectors joining adjacent pipe sections \\
\quad Val & Valve & Flow control devices mounted on pipe systems \\
\quad Pmp & Pump & Mechanical devices for fluid movement \\
\quad Str & Strainer & Filtration devices installed in pipe systems \\
\quad Elb & Elbow & Angular pipe connectors (45° or 90°) \\
\quad Tee & Tee & T-shaped pipe junction fittings \\
\quad Rdr & Reducer & Pipe diameter transition fittings \\
\bottomrule
\end{tabular}
\end{table}

\begin{figure}[H]
    \centering
    \includegraphics[width=\textwidth]{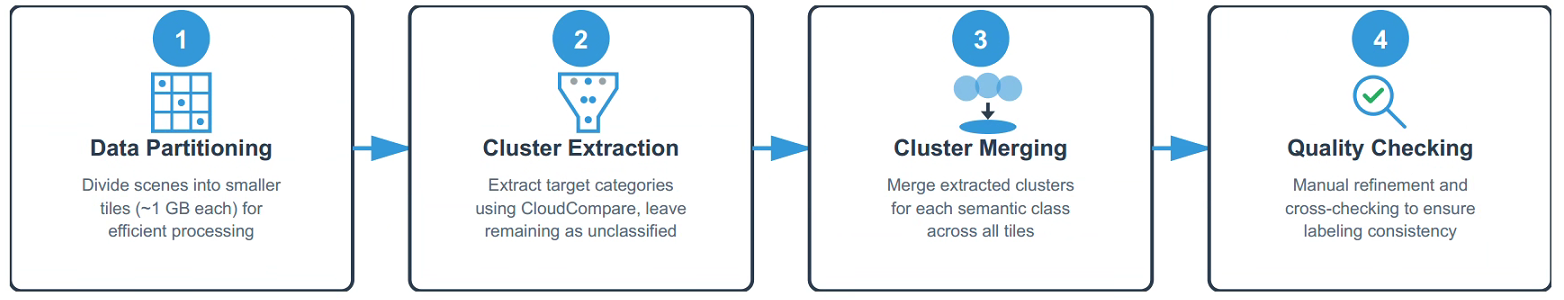}
    \caption{Four-step annotation pipeline used to label the Industrial3D dataset (612.7M points): tiling into manageable segments, per-tile cluster extraction in CloudCompare, scene-wide merging, and boundary refinement with outlier removal.}
    \label{fig:annotation_pipeline}
\end{figure}

To manage the large data volume, we developed a four-step pipeline illustrated in~\Cref{fig:annotation_pipeline}: (1) \textbf{Tiling} — we spatially divided each scene into $\sim$1\,GB tiles suitable for standard workstations; (2) \textbf{Cluster extraction} — annotators manually segmented point clusters for each of the 12 classes per tile using CloudCompare (\url{https://www.cloudcompare.org/}); (3) \textbf{Merging} — we merged per-tile clusters into scene-wide class representations; and (4) \textbf{Refinement} — we applied statistical outlier removal and manual boundary correction, followed by \rev{independent cross-checking among annotators to guarantee labelling consistency and high annotation quality} across tiles and scenes.

\begin{revblock}
The annotated scenes are room-level captures rather than object-only crops: within the accessible scanning envelope, annotators labelled the MEP and structural components visible in each registered room scene. Non-target structures such as floors, ceilings, walls, temporary clutter, and objects that could not be identified reliably were assigned to \texttt{unclassified} and excluded from the benchmark metrics. This convention separates scene completeness from class-taxonomy scope: the scans preserve room context, while the supervised task evaluates the 12 classes that can be annotated consistently across all areas.
\end{revblock}

\begin{table}[htbp]
\centering
\caption{Breakdown of the 754 person-hours spent on annotating the Industrial3D dataset by role and task. Hours = Hrs, Coordinator = Coord., Annotator = Anno.}
\label{tab:annotation_hours}
\begin{tabular}{@{}llp{6cm}c@{}}
\toprule
\textbf{Role} & \textbf{Hrs} & \textbf{Duties} & \textbf{Avg. Hrs/Tile} \\ \midrule
Anno. 1 (Coord.) & 150 & Raw data prep, preprocessing, coordination, QA & N/A \\
Anno. 2 & 192 & Labelling, QA, preprocessing & 6.3 \\
Anno. 3 & 192 & Labelling, QA, preprocessing & 6.5 \\
Anno. 4 & 50 & Labelling, QA, preprocessing & 6.8 \\
Anno. 5 & 40 & Labelling, QA, preprocessing & 7.0 \\
Refinement (Anno. 1 \& 4) & 130 & Refine clusters, data preprocessing & - \\ \midrule
\textbf{TOTAL} & \textbf{754} & & \\ \bottomrule
\end{tabular}
\end{table}

\begin{figure}[H]
    \centering
    \includegraphics[width=\textwidth]{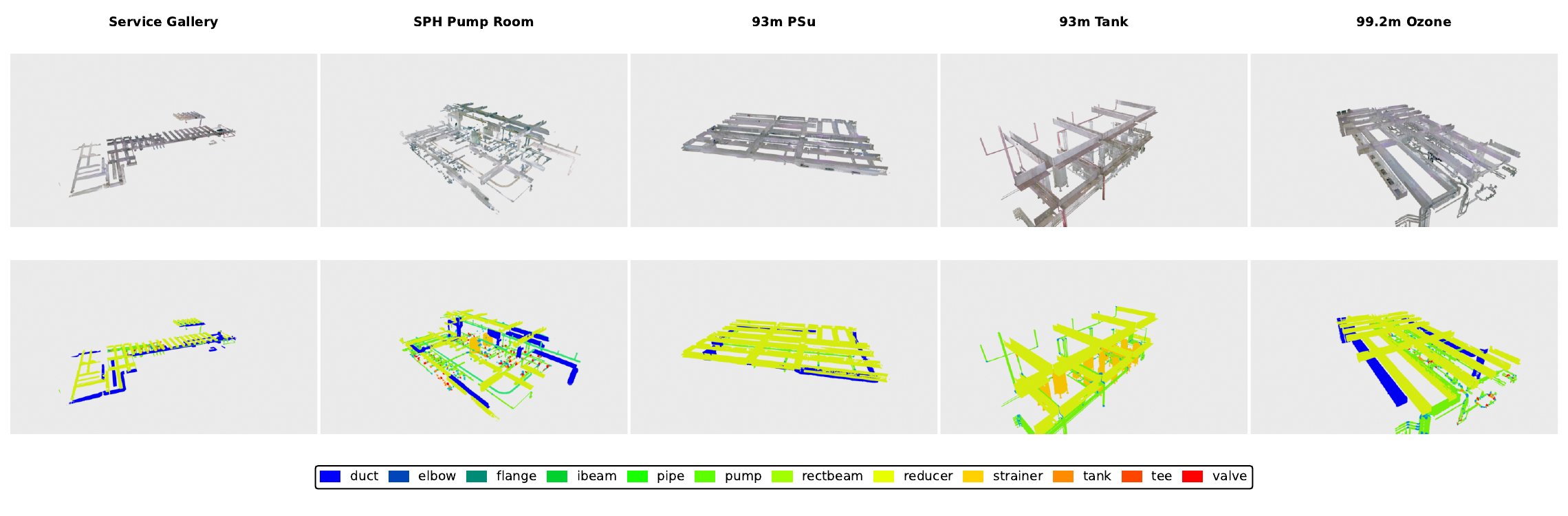}
    \caption{Five representative scenes from Industrial3D across training and test splits. Top row: raw RGB point clouds. Bottom row: semantic annotations. Areas shown: Service Gallery (Area 2, train), SPH Pump Room (Area 12, test), 93m PSU (Area 6-1, test), 93m Tank (Area 3, train), 99.2 Ozone Generation (Area 5, train). Remaining scenes in~\Cref{fig:scene_gallery_more}. \rev{All scene captures use a consistent Open3D viewpoint and colour setting; semantic panels preserve fixed class colours for comparability.}}
    \label{fig:scene_gallery}
\end{figure}

\Cref{fig:scene_gallery} presents five representative scenes from both training and test splits, showcasing annotation quality and the geometric complexity of industrial environments.

\subsection{Dataset Statistics}
\label{sec:statistics}
\begin{revblock}
The Industrial3D dataset comprises 612.7 million labelled points, each with six attributes: 3D spatial coordinates $\{X, Y, Z\}$ and RGB colour values $\{R, G, B\}$. The data span 20 room scenes grouped into 13 dataset areas across 7 operational water treatment facilities. The measured representative scenes in~\Cref{tab:room_stats} cover 11,237.06\,m\textsuperscript{2} across seven room/scene types. An Open3D audit of the released semantic PLY files gives 24,520.64\,m\textsuperscript{2} of projected labelled-geometry bounding-box footprint across the 13 areas, but this is not an architectural floor-area measurement because the public semantic files exclude \texttt{unclassified} floor, wall, ceiling, and other room-envelope points. We therefore use only the measured representative-room area as a quantitative floor-area claim.
\end{revblock}

\begin{figure}[H]
    \centering
    \includegraphics[width=\textwidth]{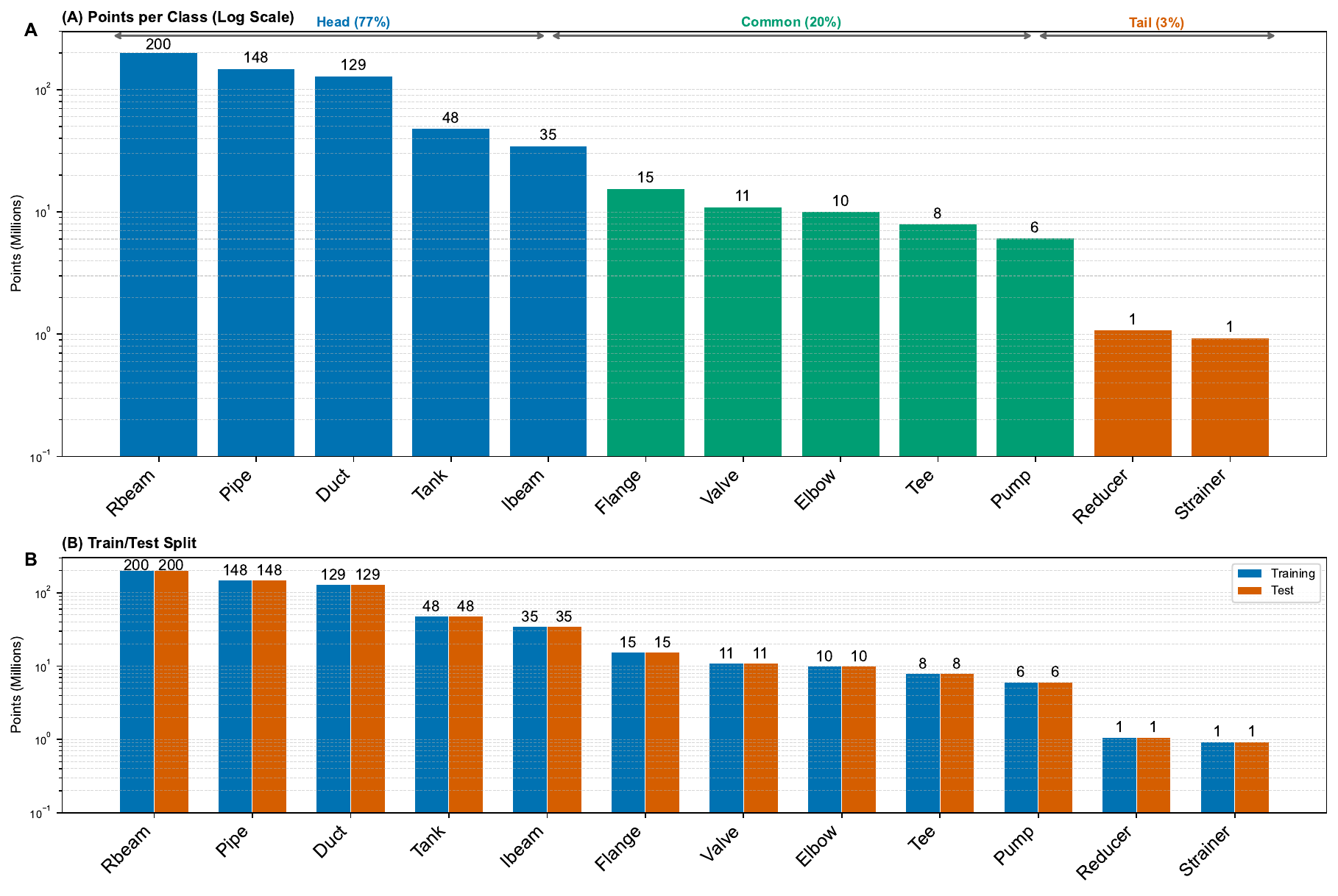}
    \caption{Class distribution and train-test split in Industrial3D. (A) Point counts per semantic class (logarithmic scale, units in millions). Classes group into three tiers: Head (77\%), Common (20\%), Tail (3\%). (B) Distribution across training (86.1\%) and test (13.9\%) sets.}
    \label{fig:industrial3d_split_distribution}
\end{figure}

\Cref{fig:industrial3d_split_distribution} illustrates the pronounced class imbalance across both the training and test splits. Dominant structural classes (rectangular beam (rbeam, 32.6\%), pipe (24.2\%), and duct (21.1\%)) collectively account for 77\% of all labelled points. In contrast, tail classes such as strainer (0.15\%) and reducer (0.18\%) are extremely rare. Per-area breakdowns are reported in~\Cref{tab:industrial3d_area_category}, showing that the spatial distribution of classes is highly non-uniform: tank points are concentrated in Areas 3 and 9, while pump and strainer points occur primarily in the Chiller (Area 1), SPH (Area 12), and WRT2 (Area 13) scenes.

\begin{figure}[H]
    \centering
    \includegraphics[width=0.95\textwidth]{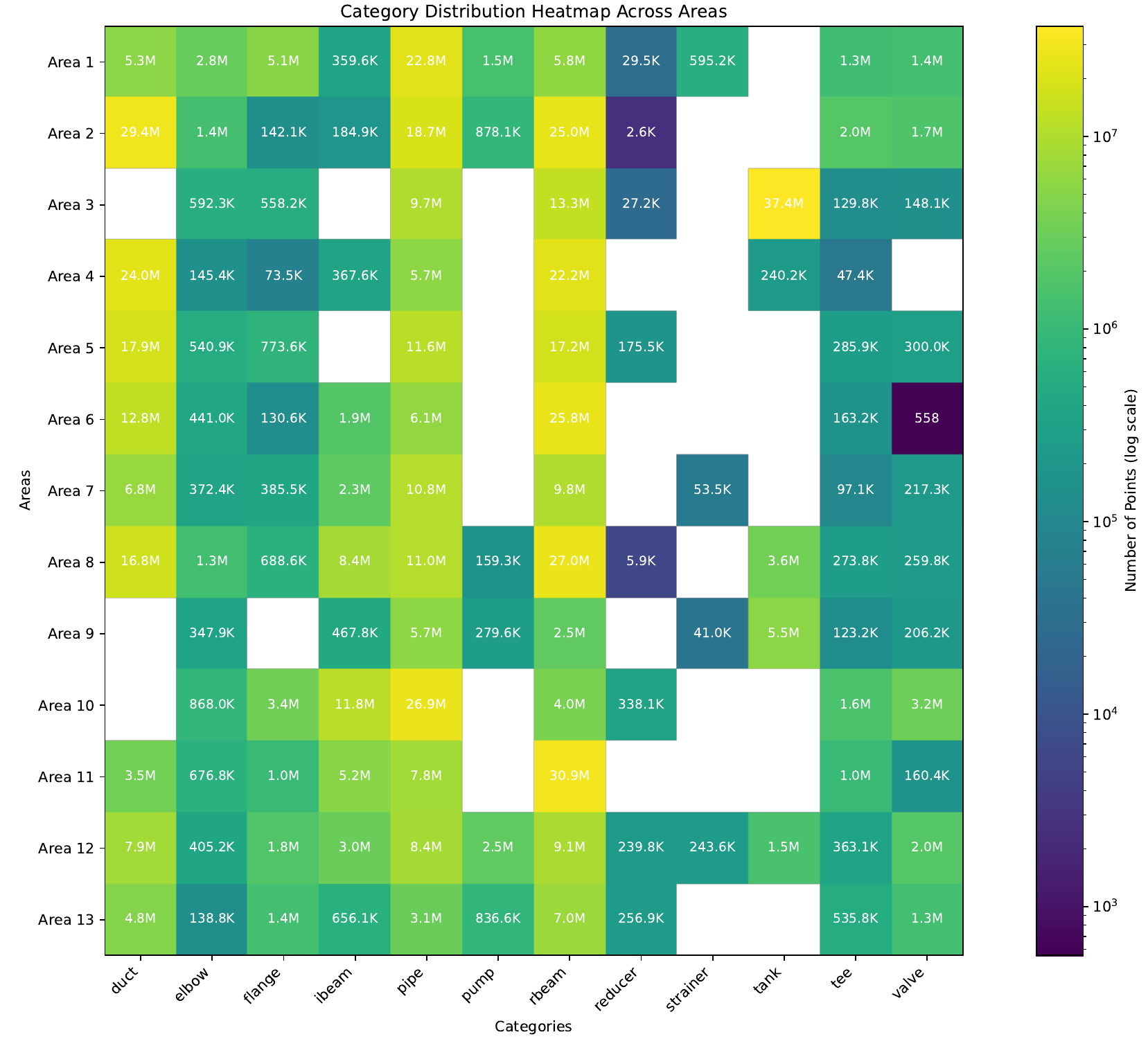}
    \caption{Semantic category distribution across 13 areas in Industrial3D (values in millions of points). Structural components (rectangular beam, pipe, duct) appear broadly distributed, while functional tail classes (pump, strainer, valve) concentrate in specific operational zones. This spatial heterogeneity reflects the functional organization of water treatment facilities.}
    \label{fig:industrial3d_distribution_heatmap}
\end{figure}

\begin{figure}[H]
    \centering
    \includegraphics[width=\textwidth]{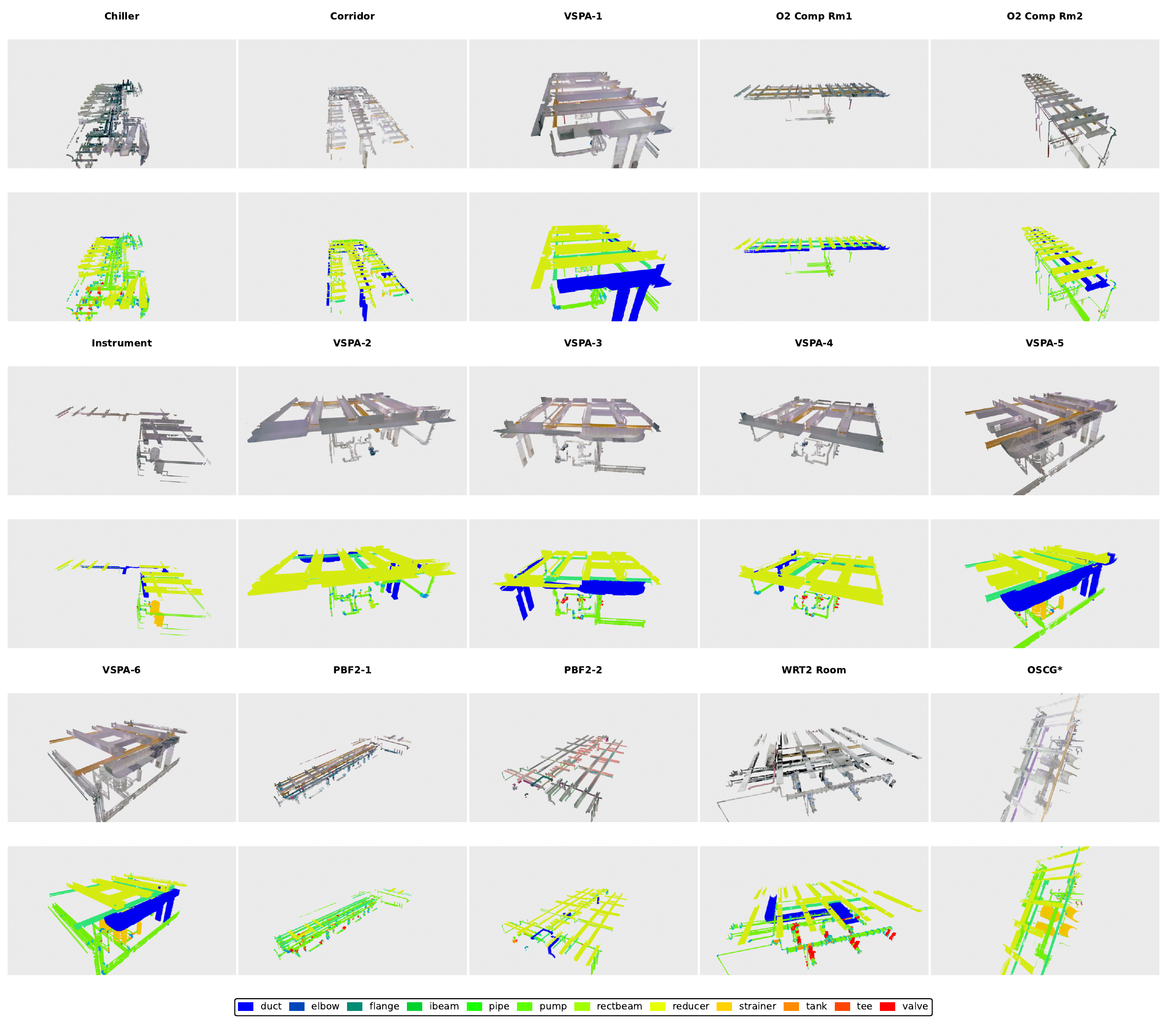}
    \caption{Complete scene gallery for Industrial3D, presenting the remaining 15 scenes. Each row pairs RGB renders (top) with semantic annotations (bottom), spanning Chiller Room (Area~1), Corridor (Area~4), 99.2 VSPA-1 (Area~6-2), Oxygen Compressor Rooms (Area~7), Instrument Room (Area~7-3), VSPA stations (Area~8), VSPA-6 (Area~8-6), PBF2 filtration rooms (Areas~10--11), WRT2 (Area~13), and OSCG (Area~9). This gallery demonstrates systematic coverage across water-treatment subsystems, from chilled water production to sludge treatment, \rev{enabling evaluation of within-domain segmentation generalisability. All scene captures use a consistent Open3D viewpoint and colour setting; semantic panels preserve fixed class colours for comparability.}}
    \label{fig:scene_gallery_more}
\end{figure}

\begin{sidewaystable}[htbp]
\centering
\caption{Point counts per area and semantic category in Industrial3D (M = million points). Test set: Areas 6 and 12 (*); Training: remaining \rev{11} areas. Inner cell values in thousands (K); row totals (rightmost column) and column totals (bottom row) in millions (M). Area abbreviations: SG=Service Gallery, O$_2$ C / I, OSCG=Ozone/Sodium Chloride Generation, PBF=Primary Biological Filter, PSU=Pumping Station Unit, SPH=Sludge Press House, VPA=Variable Pump Area, WRT=Washwater Recovery Tank, OG=Ozone Generation. Semantic class abbreviations follow~\Cref{tab:industrial3d_classes}.}
\label{tab:industrial3d_area_category}
\small
\begin{tabular}{@{}lrrrrrrrrrrrrr@{}}
\toprule
\textbf{Area (Type)} & \textbf{Dct} & \textbf{Elb} & \textbf{Flg} & \textbf{Ibm} & \textbf{Pipe} & \textbf{Pmp} & \textbf{Rbm} & \textbf{Rdr} & \textbf{Str} & \textbf{Tank} & \textbf{Tee} & \textbf{Val} & \textbf{Total} \\ \midrule
1 (Chiller)                      & 5,311  & 2,770 & 5,150  & 360    & 22,800 & 1,478 & 5,821  & 29  & 595 & 0      & 1,252 & 1,389 & \textbf{47.0 M} \\
\rev{2 (SG)}              & 29,443 & 1,432 & 142    & 185    & 18,740 & 878   & 25,016 & 3   & 0   & 0      & 2,011 & 1,743 & \textbf{\rev{79.6 M}} \\
3 (Tank)                        & 0      & 592   & 558    & 0      & 9,738  & 0     & 13,340 & 27  & 0   & 37,392 & 130   & 148   & \textbf{61.9 M} \\
4 (Corridor)                    & 24,007 & 145   & 74     & 368    & 5,668  & 0     & 22,232 & 0   & 0   & 240    & 47    & 0     & \textbf{52.8 M} \\
5 (OG)            & 17,920 & 541   & 774    & 0      & 11,609 & 0     & 17,204 & 176 & 0   & 0      & 286   & 300   & \textbf{48.8 M} \\
6* (PSU / VPA) & 12,819 & 441   & 131    & 1,907  & 6,122  & 0     & 25,809 & 0   & 0   & 0      & 163   & 1     & \textbf{47.4 M} \\
7 (O$_2$ C/I) & 6,775  & 372   & 385    & 2,325  & 10,764 & 0     & 9,836  & 0   & 53  & 0      & 97    & 217   & \textbf{30.8 M} \\
8 (VPA)                     & 16,824 & 1,313 & 689    & 8,443  & 10,957 & 159   & 26,961 & 6   & 0   & 3,550  & 274   & 260   & \textbf{69.4 M} \\
\rev{9 (OSCG)}                         & 0      & 348   & 0      & 468    & 5,713  & 280   & 2,488  & 0   & 41  & 5,456  & 123   & 206   & \textbf{\rev{15.1 M}} \\
10 (PBF-1)                      & 0      & 868   & 3,369  & 11,811 & 26,863 & 0     & 4,038  & 338 & 0   & 0      & 1,629 & 3,182 & \textbf{52.1 M} \\
11 (PBF-2)                      & 3,459  & 677   & 1,029  & 5,167  & 7,843  & 0     & 30,862 & 0   & 0   & 0      & 1,017 & 160   & \textbf{50.2 M} \\
12* (SPH)                         & 7,943  & 405   & 1,824  & 3,023  & 8,409  & 2,455 & 9,056  & 240 & 244 & 1,523  & 363   & 2,027 & \textbf{37.5 M} \\
13 (WRT)                        & 4,785  & 139   & 1,365  & 656    & 3,053  & 837   & 7,041  & 257 & 0   & 0      & 536   & 1,332 & \textbf{20.0 M} \\ \midrule
\textbf{Total (M)}                 & \textbf{129.3} & \textbf{10.1} & \textbf{15.5} & \textbf{34.7} & \textbf{148.3} & \textbf{6.1} & \textbf{199.7} & \textbf{1.1} & \textbf{0.9} & \textbf{48.2} & \textbf{7.9} & \textbf{11.0} & \textbf{612.7} \\ \bottomrule
\end{tabular}
\end{sidewaystable}

\Cref{tab:dataset_split} summarises the area-based training and test split, following the S3DIS evaluation protocol~\citep{armeni2016}. Areas 6 and 12 (93m PSU/99.2 VSPA-1 and SPH) form the held-out test set (84.9M points, 13.9\%); \rev{the remaining 11 areas, including Area 9 (OSCG)}, constitute the training set (527.8M points, 86.1\%). Complete per-area per-class statistics are reported in~\Cref{tab:industrial3d_area_category}.

Industrial3D embodies what this paper terms the \textit{dual crisis} in industrial point cloud segmentation: the simultaneous occurrence of extreme statistical rarity and geometric ambiguity. The 215:1 head-to-tail imbalance (Rbm:Str) is 3.5$\times$ more severe than S3DIS (62:1), and over 80\% of tail-class points share cylindrical geometric primitives with the dominant pipe class. Tail classes fall into two subtypes: \textit{composite-tail} classes (Flg, Val, Pmp, Str), whose parts share primitives with head classes, and \textit{primitive-similarity} tail classes (Elb, Tee, Rdr), which are locally indistinguishable from pipes due to shared cylindrical structure. A detailed analysis with benchmark-based evidence is presented in~\Cref{sec:challenges}; proposed solutions to this challenge are described in the companion methods paper~\citep{Yin2026arXiv}.

Four characteristics distinguish Industrial3D from architectural benchmarks: (1) dense, interconnected component arrangements with frequent mutual occlusions; (2) elongated structures (pipes, beams) spanning large spatial extents; (3) functional components (valves, pumps) embedded deep within pipe networks; and (4) limited colour contrast among metallic components, reducing the discriminative utility of RGB features relative to geometric cues.

\subsection{Dataset Comparison}
\label{sec:comparison}
\Cref{tab:related_work_comparison} compares Industrial3D against representative large-scale point cloud datasets. Industrial3D offers \rev{four} key distinctions. First, \textbf{scale}: at 612.7M points, it represents a 6.6$\times$ increase over the closest MEP dataset (\citealt{jing2024mep}, 92.1M points) and a 4$\times$ increase over construction datasets. \rev{Second, \textbf{within-domain diversity}: the dataset covers 20 room scenes, 13 dataset areas, and multiple water-treatment subsystems across 7 operational facilities, allowing methods to be evaluated on held-out areas rather than repeated views of a single room type. Third}, \textbf{domain specificity}: unlike MEP datasets targeting common buildings~\citep{jing2024mep} or general construction sites, Industrial3D focuses exclusively on operational water treatment facilities. Their dense equipment arrangements, functional interdependencies, and metallic appearance create qualitatively different challenges. \rev{Fourth, \textbf{challenge severity}: the 215:1 class imbalance ratio is the most extreme among the compared benchmarks, compounded by the geometric ambiguity described above. Together, these properties make Industrial3D a uniquely demanding benchmark for robust water-treatment MEP scene understanding}.

Industrial3D also extends our prior work PSNet5~\citep{yin2021}, which introduced the first industrial MEP dataset with 80M points and 5 classes. The 7.7$\times$ scale expansion and addition of 8 new functional classes (flange, valve, pump, strainer, duct, tank, ibeam, reducer) position Industrial3D as a substantially enhanced successor to PSNet5, enabling research directions that were previously infeasible---including 0.1\% sparse-supervision experiments, cross-area generalisation across 13 areas, and \rev{downstream MEP modelling protocols requiring the separation of functionally distinct but geometrically similar components}.

\section{Benchmark}
\label{sec:benchmark}

Industrial3D enables the first cross-paradigm evaluation on real-world industrial point clouds. We benchmark nine representative methods across four learning paradigms (fully supervised, weakly supervised, unsupervised, and foundation models) to measure how current approaches transfer from standard benchmarks to industrial MEP scenes. This section defines the dataset split, introduces the baselines, describes the metrics, and reports the quantitative results.
\subsection{Dataset Split}

Following the S3DIS evaluation protocol~\citep{armeni2016}, we partition Industrial3D into training and test sets using entire areas to prevent data leakage. \rev{The training set comprises 11 areas (527.8M points, 86.1\%) spanning diverse functional zones including chiller plants, service galleries, ozone generation rooms, OSCG, VSPA rooms, washwater recovery, and biological filtration areas. The held-out test set contains 2 areas (84.9M points, 13.9\%): the Pumping Station Unit and VSPA-1 rooms grouped as Area~6}, and the Sludge Press House (Area~12). These two areas were selected because they represent spatially distinct \rev{water-treatment subsystems with complementary characteristics---Area~6 contains the 93m PSU and 99.2 VSPA-1 rooms with dense piping arrangements, while Area~12 contains diverse mechanical equipment---ensuring evaluation on unseen area types within the same water-treatment domain}. \Cref{tab:dataset_split} provides the complete area-wise breakdown, and~\Cref{tab:industrial3d_area_category} reports the per-area per-class statistics.

\begin{table}[htbp]
\centering
\caption{\rev{Industrial3D dataset split statistics following the S3DIS area-based protocol (M = million points). Training: 11 areas (527.8M, 86.1\%), including Area 9 (OSCG)}. Test: 2 areas, 6+12 (84.9M, 13.9\%). Area abbreviations: SG=Service Gallery, OG=Ozone Generation, PSU=Pumping Station Unit, VPA=Variable Pump Area, O$_2$ C/I=Oxygen Compressor/Instrument, OSCG=Ozone/Sodium Chloride Generation, PBF=Primary Biological Filter, SPH=Sludge Press House, WRT=Washwater Recovery Tank. In area type, $/$ denotes multiple room types in one area.}
\label{tab:dataset_split}
\begin{tabular}{@{}lrlr@{}}
\toprule
\textbf{Split} & \textbf{Index} & \textbf{Area type} & \textbf{Points (M)} \\ \midrule
\textit{Training} & 1 & Chiller & 47.0M \\
& 2 & SG & 79.6M \\
& 3 & Tank & 61.9M \\
& 4 & Corridor & 52.8M \\
& 5 & OG & 48.8M \\
& 7 & O$_2$ C/I & 30.8M \\
& 8 & VPA & 69.4M \\
& \rev{9} & \rev{OSCG} & \rev{15.1M} \\
& 10 & PBF-1 & 52.1M \\
& 11 & PBF-2 & 50.2M \\
& 13 & WRT & 20.0M \\ \midrule
\textit{Test} & 6 & PSU / VPA & 47.4M \\
& 12 & SPH & 37.5M \\ \midrule
\textbf{Total} & & & \textbf{612.7M} \\ \bottomrule
\end{tabular}
\end{table}

The training set exhibits extreme class imbalance characteristic of industrial MEP scenes, directly reflecting the statistical rarity component of the dual crisis. Three dominant classes (rectangular beam, duct, and pipe) account for 77\% of training points, while the seven rarest classes each comprise less than 3\%. This 215:1 head-to-tail ratio (rectangular beam:strainer) presents substantial challenges for long-tailed learning. \Cref{sec:results} demonstrates how this dual crisis affects methods across all learning paradigms.

\subsection{Representative Baselines}

We selected representative methods across four learning paradigms using three criteria: (1) state-of-the-art performance on architectural benchmarks (S3DIS, ScanNet), (2) relevance to industrial challenges identified in Section~\ref{sec:challenges}, and (3) diversity of architectural approaches to enable meaningful cross-paradigm comparison. This selection facilitates comprehensive evaluation of how different learning strategies handle the Industrial3D dataset.

\textbf{Implementation note.} To enable unified and reproducible evaluation, we reproduced methods lacking official PyTorch implementations in PyTorch. Most notably, our SQN PyTorch implementation achieves 60.55\% mIoU on S3DIS Area 5 with 0.1\% labels, within 0.85\% of the original TensorFlow results (61.4\%), validating implementation fidelity (see Appendix~\ref{sec:sqn_reproduction}). This TensorFlow-to-PyTorch reimplementation effort ensures that all nine benchmarked methods operate under a single consistent training framework, eliminating confounds from framework-specific behaviors.

\textbf{Fully-supervised baselines} represent the current state of the art in point cloud segmentation:
\begin{itemize}
    \item \textbf{KPConv}~\citep{thomas2019}: Kernel point convolution networks with rigid/deformable kernel points for learning local geometries in spherical neighbourhoods; achieves strong performance on S3DIS through explicit geometric modelling.
    \item \textbf{PosPool}~\citep{liu2020pospool}: Position-aware pooling for hierarchical feature learning across multi-scale point representations; designed to handle the varying point density common in TLS data.
    \item \textbf{RandLA-Net}~\citep{hu2020randlanet}: Efficient random sampling and local feature aggregation for large-scale scan processing; provides a lightweight baseline for industrial applications.
    \item \textbf{ResPointNet++}~\citep{yin2021}: Residual connections integrated into hierarchical feature learning, specifically designed for industrial point clouds; our prior work establishing the industrial MEP segmentation task.
    \item \textbf{PTv3}~\citep{wu2024point}: Transformer architecture with serialised point ordering for efficient localised attention, benchmarking the attention-based method family.
    \item \textbf{Boundary-CB}~\citep{Yin2026arXiv}: Class-balanced loss with boundary-aware features; explicitly addresses the 215:1 head-to-tail imbalance through frequency re-weighting and geometric context modelling.
\end{itemize}

\textbf{Weakly-supervised baselines} evaluate label-efficient learning critical for industrial applications where annotation costs are prohibitive:
\begin{itemize}
    \item \textbf{SQN}~\citep{hu2022sqn}: Sparse query network exploiting semantic homogeneity in local neighbourhoods; enables training from as few as 0.01--0.1\% labelled points by propagating supervision through larger receptive fields than conventional U-Net architectures.
\end{itemize}

\textbf{Unsupervised baselines} assess representation learning without any labels:
\begin{itemize}
    \item \textbf{GrowSP}~\citep{zhang2023growsp}: Unsupervised semantic feature learning via low-level geometric clustering with progressively growing superpoints.
\end{itemize}

\textbf{Foundation models} evaluate zero-shot and few-shot transfer from large-scale pre-training:
\begin{itemize}
    \item \textbf{Point-SAM}~\citep{zhou2025pointsam}: Segment Anything model adapted to 3D point clouds; enables zero-shot segmentation through prompt engineering and few-shot domain adaptation with minimal labelled data.
\end{itemize}

\subsection{Metrics and Experimental Setup}
\subsubsection{Evaluation Metrics}
\label{sec:benchmark_metrics}
We employ a layered suite of metrics tailored to each learning paradigm. Fully supervised and weakly supervised methods are assessed with standard per-point semantic metrics (mIoU, OA) and long-tailed partitioned metrics. Unsupervised methods require an additional cluster-to-class assignment step via Hungarian matching before the same metrics apply. Foundation models are evaluated with IoU@k, the canonical metric for promptable segmentation, adapted to class-level semantic evaluation on Industrial3D.

\textbf{Standard Metrics (Fully-Supervised and Weakly-Supervised).}
For a dataset with $C$ classes, let $TP_c$, $FP_c$, $FN_c$ denote the true positives, false positives, and false negatives for class $c$, respectively. The per-class Intersection-over-Union is:
\begin{equation}
\label{eq:iou}
\text{IoU}_c = \frac{TP_c}{TP_c + FP_c + FN_c}
\end{equation}
The mean IoU (mIoU) averages uniformly across all $C$ classes, giving equal weight to rare and common classes regardless of their point-count share:
\begin{equation}
\label{eq:miou}
\text{mIoU} = \frac{1}{C} \sum_{c=1}^{C} \text{IoU}_c
\end{equation}
Overall accuracy (OA) measures the fraction of correctly classified points:
\begin{equation}
\label{eq:oa}
\text{OA} = \frac{\sum_{c=1}^{C} TP_c}{\sum_{c=1}^{C} (TP_c + FN_c)}
\end{equation}
We report mIoU and OA for methods producing per-point semantic labels directly: fully supervised (KPConv, PosPool, RandLA-Net, ResPointNet++, PTv3, Boundary-CB) and weakly supervised (SQN) methods. While mIoU mitigates the frequency bias of OA, both metrics can still mask catastrophic failure on rare classes when high-frequency head classes dominate.

\textbf{Long-Tailed Metrics.}
Industrial3D's 215:1 head-to-tail class imbalance demands metrics that expose per-group performance explicitly. Following Boundary-CB~\citep{Yin2026arXiv}, we partition the 12 classes into three groups: $\mathcal{C}_{\text{head}}$ (the 3 most frequent classes: RectBeam, Pipe, Duct; collectively 77\% of points), $\mathcal{C}_{\text{common}}$ (2 moderately frequent classes: IBeam, Tank), and $\mathcal{C}_{\text{tail}}$ (the 7 rarest classes, each $<$3\% of points). Group-wise mIoUs are:
\begin{equation}
\label{eq:head_miou}
\text{mIoU}_{\text{head}} = \frac{1}{|\mathcal{C}_{\text{head}}|} \sum_{c \in \mathcal{C}_{\text{head}}} \text{IoU}_c
\end{equation}
\begin{equation}
\label{eq:common_miou}
\text{mIoU}_{\text{common}} = \frac{1}{|\mathcal{C}_{\text{common}}|} \sum_{c \in \mathcal{C}_{\text{common}}} \text{IoU}_c
\end{equation}
\begin{equation}
\label{eq:tail_miou}
\text{mIoU}_{\text{tail}} = \frac{1}{|\mathcal{C}_{\text{tail}}|} \sum_{c \in \mathcal{C}_{\text{tail}}} \text{IoU}_c
\end{equation}
The Harmonic Mean IoU (H-IoU) captures the head--tail trade-off in a single scalar:
\begin{equation}
\label{eq:hiou}
\text{H-IoU} = \frac{2 \cdot \text{mIoU}_{\text{head}} \cdot \text{mIoU}_{\text{tail}}}{\text{mIoU}_{\text{head}} + \text{mIoU}_{\text{tail}}}
\end{equation}
A high H-IoU indicates that tail-class gains have been achieved without degrading head-class accuracy. This balance is critical for Scan-to-BIM applications: missing a rare valve or strainer can invalidate an entire MEP system model.

\textbf{Unsupervised Evaluation with Hungarian Matching.}
Unsupervised methods such as GrowSP~\citep{zhang2023growsp} produce cluster indices rather than semantic labels. Before computing mIoU or OA, predicted cluster assignments must be mapped to ground-truth semantic classes via \textit{Hungarian matching} (linear sum assignment). This one-time matching is performed globally on the test set and does not use any labelled training data. This gives unsupervised methods the benefit of optimal post-hoc label alignment, making the resulting mIoU an upper bound on true unsupervised performance.

\textbf{Foundation Model Metrics: IoU@k and Industrial3D Adaptations.}
Point-SAM~\citep{zhou2025pointsam} is evaluated natively using IoU@k (interactive segmentation quality after $k$ point prompts). Since Industrial3D requires class-level semantic segmentation rather than instance segmentation, we adapt this framework with two protocols: (1) \textit{Oracle mIoU} uses ground-truth masks to guide prompt selection over $T=5$ iterations, providing an upper bound independent of prompt quality; (2) \textit{One-vs-Rest mIoU} samples five positive and negative prompts per class at random, approximating realistic zero-shot deployment. The oracle--one-vs-rest gap quantifies prompt sensitivity versus model capacity limitations.

\subsubsection{Experimental Setup}
We conducted all experiments using official PyTorch implementations where available. For methods without official PyTorch releases, we reproduced the code and verified performance parity with original reported results. Specifically, our SQN implementation was validated on S3DIS Area 5 (achieving 60.55\% mIoU with 0.1\% labels, within 0.85\% of the reported 61.4\%), as detailed in Appendix~\ref{sec:sqn_reproduction}. Hyperparameters follow each method's original publication; given Industrial3D's structural similarity to S3DIS (both are indoor point cloud datasets with similar spatial resolution), we use S3DIS hyperparameters as baselines with minor adjustments for the increased scale (612.7M vs. 273.6M points) and class count (12 vs. 13). All training was performed on one or two NVIDIA RTX 3090 GPUs with 24GB memory.

Our GitHub repository (\url{https://github.com/pointcloudyc/Industrial3D}) provides complete reproducibility: configuration files, training scripts, pre-trained model weights, and detailed documentation for reproducing baseline experiments reported in this benchmark.

\subsection{Results}
\label{sec:results}
We present the quantitative and qualitative results of the 9 baseline methods on the Industrial3D test set.

\subsubsection{Fully-Supervised Methods}

Fully supervised methods achieve strong overall performance (55.74\% mIoU) but exhibit severe head-tail disparity that reveals the dual crisis in action: even with boundary-aware loss re-weighting, a 58.55 pp gap persists between head classes (88.14\%) and tail classes (29.59\%). \Cref{tab:benchmark_results_fs} presents the detailed mIoU and per-class IoU results.
\begin{table}[htbp]
    \centering
    \caption{Fully-supervised benchmark results on Industrial3D test set (mIoU, per-class IoU). CB indicates class-balanced weighting loss. Best results per column are in bold.}
    \label{tab:benchmark_results_fs}
    \resizebox{\textwidth}{!}{
    \begin{tabular}{lccccccccccccc}
    \hline
    \textbf{Method} & \textbf{mIoU} & \textbf{Dct} & \textbf{Elb} & \textbf{Flg} & \textbf{Ibm} & \textbf{Pipe} & \textbf{Pmp} & \textbf{Rbm} & \textbf{Rdr} & \textbf{Str} & \textbf{Tank} & \textbf{Tee} & \textbf{Val} \\
    \hline
    RandLA-Net~\citep{hu2020randlanet} & 39.83 & 83.37 & 18.06 & 23.92 & 81.14 & 68.97 & 0.98 & 91.25 & 0.00 & 0.00 & 71.89 & 3.03 & 35.36 \\
    PTv3~\citep{wu2024point} & 41.90 & \textbf{97.39} & \textbf{55.40} & 33.35 & \textbf{99.71} & \textbf{91.80} & 0.00 & \textbf{98.53} & 0.00 & 0.00 & 0.00 & 26.61 & 0.00 \\
    ResPointNet++~\citep{yin2021} & 52.48 & 90.73 & 40.77 & 45.23 & 98.66 & 76.33 & 43.08 & 94.79 & 0.00 & 0.00 & 98.99 & 3.10 & 38.06 \\
    KPConv~\citep{thomas2019} & 53.65 & 91.35 & 45.44 & 50.95 & 98.35 & 75.96 & 42.60 & 95.04 & 0.00 & 0.00 & 97.72 & 7.72 & \textbf{38.69} \\
    PosPool~\citep{liu2020pospool} & 53.18 & 90.25 & 32.93 & 42.88 & 98.93 & \textbf{76.35} & 31.27 & 94.18 & \textbf{34.48} & 0.00 & 95.59 & 9.17 & 32.14 \\
    Boundary-CB~\citep{Yin2026arXiv} & \textbf{55.74} & 91.08 & 43.23 & 47.85 & 98.51 & 78.32 & 44.37 & 95.02 & 21.12 & 0.00 & \textbf{98.76} & 3.24 & \textbf{47.32} \\
    \hline
    \end{tabular}
    }
    \vspace{0.1in}
\end{table}

\textbf{Overall performance.} Boundary-CB achieves the best mIoU (55.74\%) by explicitly addressing the long-tailed challenge, followed closely by KPConv (53.65\%), PosPool (53.18\%), and the baseline ResPointNet++ (52.48\%). PTv3 achieves only 41.90\% mIoU despite its state-of-the-art performance on indoor benchmarks, performing only marginally above RandLA-Net (39.83\%). This 15.91-pp gap from the best method confirms that strong architectural performance on indoor datasets does not transfer directly to the extreme imbalance and geometric ambiguity of industrial scenes.

\textbf{Head-versus-tail disparity.} All methods exhibit severe performance imbalance reflecting the dual crisis. Large structural components achieve consistently high IoU---Rbeam (91--99\%), Duct (90--97\%), Pipe (68--92\%), Tank (71--99\%)---while tail classes suffer catastrophic failure: Strainer (0\% IoU across most methods), Reducer (0--35\%), and Tee (3--27\%) are nearly unrecognisable. PTv3 illustrates this extreme clearly: it achieves outstanding head-class performance (Duct: 97.39\%, Rbeam: 98.53\%, Ibeam: 99.71\%) but scores 0\% IoU on five tail classes. Strainer failure is attributable to three compounding factors: insufficient training representation (0.6\% of points), cylindrical geometry shared with Pipe, and small physical size. For Reducer and Tee, geometric ambiguity is the primary driver; cylindrical primitive overlap with Pipe creates systematic confusion that the class-balanced variant only partially mitigates (Reducer: 0\% $\to$ 21.12\% with Boundary-CB).

\textbf{Architectural insights.} Boundary-CB improves tail-class mIoU by 5.27~pp over the ResPointNet++ baseline through combined frequency re-weighting and boundary-aware feature modelling, yet a 58.55-pp head-tail gap (88.14\% vs.\ 29.59\%) persists. The performance ceiling at 55.74\% mIoU---more than 15~pp below state-of-the-art scores on S3DIS ($>$70\%)---confirms that Industrial3D poses a qualitatively harder problem. With this ceiling established, we now assess whether label-efficient methods can approach comparable performance with substantially less annotation.

\subsubsection{Weakly-Supervised Methods}

Fully supervised results establish a 55.74\% ceiling but require 612.7M fully annotated points (754 person-hours). We now evaluate SQN~\citep{hu2022sqn}, which uses semantic query propagation to learn from only 0.01--0.1\% of labelled points. mIoU and OA are evaluated identically to the fully supervised setting, since SQN produces per-point semantic labels.

\Cref{tab:industrial3d_wsl} presents results with RandLA-Net as the backbone at 0.01\% and 0.1\% label ratios. SQN with 0.1\% labels achieves 44.29\% mIoU, exceeding fully supervised RandLA-Net (39.83\%) within the same architectural family. This counter-intuitive result is explained by sparse supervision acting as an implicit regulariser: observing only $M{=}40$ labelled points per batch prevents memorisation, shrinking the train--validation gap from 45.6~pp to 37.0~pp. This comparison is specific to the RandLA-Net backbone; pairing SQN with stronger backbones such as ResPointNet++ or PosPool or PTv3 remains future work.

\begin{table}[htbp]
    \centering
    \caption{Weakly-supervised SQN performance on Industrial3D using RandLA-Net backbone and vote-based evaluation (test\_smooth=0.95). Best results per column are in \textbf{bold}.}
    \label{tab:industrial3d_wsl}
    \resizebox{\textwidth}{!}{
    \begin{tabular}{lccccccccccccccc}
    \hline
    \textbf{Method} & \textbf{Labels} & \textbf{OA} & \textbf{mIoU} & \textbf{Dct} & \textbf{Elb} & \textbf{Flg} & \textbf{Ibm} & \textbf{Pipe} & \textbf{Pmp} & \textbf{Rbm} & \textbf{Rdr} & \textbf{Str} & \textbf{Tank} & \textbf{Tee} & \textbf{Val} \\
    \hline
    RandLA-Net~\citep{hu2020randlanet} & 100\% & 85.47 & 39.83 & 83.37 & 18.06 & 23.92 & 81.14 & 68.97 & 0.98 & 91.25 & 0.00 & 0.00 & 71.89 & 3.03 & 35.36 \\
    SQN~\citep{hu2022sqn} & 0.01\% & 82.31 & 33.16 & 82.46 & 0.00 & 11.06 & 65.65 & 59.66 & \textbf{8.12} & 89.55 & 0.00 & 0.00 & 74.68 & 0.00 & 6.70 \\
    SQN~\citep{hu2022sqn} & 0.1\% & \textbf{88.78} & \textbf{44.29} & \textbf{89.77} & \textbf{24.17} & \textbf{28.60} & \textbf{98.64} & \textbf{71.90} & 2.72 & \textbf{94.59} & 0.00 & 0.00 & \textbf{87.83} & 1.83 & 31.46 \\
    \hline
    \end{tabular}
    }
\end{table}

A notable limitation emerges at semantic boundaries. SQN's nearest-neighbour interpolation assumes nearby points share similar semantics, which fails when different objects abut. This is especially problematic for industrial MEP scenes where functionally distinct components (e.g., flanges connecting different pipe systems) are spatially adjacent. Our analysis reveals that tail classes exhibiting geometric ambiguity (elbow, tee, reducer) suffer most from this limitation; these components share cylindrical primitives with head-class pipes, and their local neighbourhoods often contain points from multiple semantic categories.

At 0.01\,\% labels (33.16\% mIoU), rare classes (Strainer, Reducer, Tee) consistently achieve 0\% IoU due to insufficient label coverage, confirming that the dual crisis extends to the weakly supervised regime: even a handful of labelled points cannot overcome the combined effect of statistical rarity and geometric ambiguity. These findings raise the question of whether removing labels entirely could still yield useful representations.

\subsubsection{Unsupervised Methods}

Unsupervised methods eliminate annotation requirements entirely by relying solely on point cloud geometry. We evaluated GrowSP~\citep{zhang2023growsp}, which learns semantic groupings through unsupervised primitive clustering and progressive superpoint growing. Because GrowSP produces cluster indices rather than semantic labels, we apply Hungarian matching to optimally assign predicted clusters to ground-truth classes before computing mIoU and OA; the reported numbers thus represent an upper bound on true unsupervised performance. \Cref{tab:benchmark_results_usl} presents the results.

\begin{table}[htbp]
    \centering
    \caption{Unsupervised GrowSP performance on Industrial3D test set via Hungarian matching. Best results in \textbf{bold}.}
    \label{tab:benchmark_results_usl}
    \resizebox{\textwidth}{!}{
    \begin{tabular}{lcccccccccccccc}
    \hline
    \textbf{Method} & \textbf{OA} & \textbf{mIoU} & \textbf{Dct} & \textbf{Elb} & \textbf{Flg} & \textbf{Ibm} & \textbf{Pipe} & \textbf{Pmp} & \textbf{Rbm} & \textbf{Rdr} & \textbf{Str} & \textbf{Tank} & \textbf{Tee} & \textbf{Val} \\
    \hline
    GrowSP~\citep{zhang2023growsp} & 41.81 & 11.73 & 16.37 & 6.66 & 6.18 & 17.56 & 55.02 & 0.00 & 39.00 & 0.00 & 0.00 & 0.00 & 0.00 & 0.00 \\
    \hline
    \end{tabular}
    }
    \vspace{0.1in}
\end{table}

After Hungarian matching, GrowSP achieves 11.73\% mIoU (below zero-shot Point-SAM at 15.79\% mIoU, weakly supervised SQN at 0.1\% labels at 44.29\% mIoU, and fully supervised Boundary-CB at 55.74\% mIoU). This cross-paradigm performance hierarchy reveals that unsupervised geometry-only representation learning is the most constrained paradigm on Industrial3D, unable to benefit from large-scale pretraining or sparse annotation signals.

The per-class breakdown reveals systematic failure on fine-grained MEP components: GrowSP achieves 0\% IoU on six tail classes (Pump, Reducer, Strainer, Tank, Tee, Valve) and single-digit performance on Elbow (6.66\%) and Flange (6.18\%). Only Pipe (55.02\%) and RectBeam (39.00\%) yield meaningful results. The geometric shortcut problem~\citep{wu2025sonata}, whereby models collapse to low-level spatial features rather than semantic representations, is particularly acute here: 86\% of tail-class points share cylindrical primitives with head-class pipes, so contrastive learning on geometry alone cannot distinguish them. The Hungarian matching step allocates the large ``cylinder'' cluster to Pipe, leaving tail classes with insufficient cluster coverage and near-zero IoU.

\textbf{Cross-paradigm comparison.} GrowSP (11.73\% mIoU) anchors the lower bound. Adding just 0.1\% labelled data via SQN raises mIoU to 44.29\%, a 32.56 pp gain at negligible annotation cost. Zero-shot Point-SAM (15.79\%) falls between these two extremes, confirming that large-scale pretraining on diverse 3D data delivers modest but real benefit over pure geometry-based clustering, while still falling far short of any form of domain supervision. This establishes a clear performance hierarchy: unsupervised < foundation models < weakly supervised < fully supervised, with the 39.95 pp gap between zero-shot foundation models and fully supervised methods quantifying the domain transfer distance that must be overcome.

\subsubsection{Foundation Models: Zero-Shot Evaluation and Domain Adaptation}

Point-SAM~\citep{zhou2025pointsam} is a promptable 3D segmentation foundation model whose native evaluation metric is IoU@k (segmentation quality after $k$ iterative point prompts), measuring performance on individual object instances. On its original benchmarks, Point-SAM achieves IoU@1 of 47.6--63.6\% and IoU@5 of 74.2--86.2\%, demonstrating strong interactive segmentation. To assess zero-shot transfer to industrial MEP semantics, we adapt this framework to class-level semantic evaluation using the oracle and one-vs-rest protocols defined in~\Cref{sec:benchmark_metrics}. Both protocols omit any Industrial3D training data; the oracle variant is analogous to IoU@5 with ground-truth-anchored prompts (upper bound), while one-vs-rest approximates a realistic user interaction scenario with five randomly sampled positive and negative prompts per class. \Cref{tab:benchmark_results_fm} reports per-class IoU alongside head/tail group averages.

\begin{table}[htbp]
    \centering
    \caption{Foundation model zero-shot evaluation on Industrial3D test set using Point-SAM~\citep{zhou2025pointsam}. \textit{Oracle}: GT binary masks guide prompt selection over $T{=}5$ iterations (upper bound). \textit{One-vs-Rest}: 5 positive and 5 negative prompts per class (realistic zero-shot). Head: RectBeam, Duct, Pipe. Tail: Elbow, Flange, Pump, Reducer, Strainer, Tee, Valve. Best results \textbf{bold}.}
    \label{tab:benchmark_results_fm}
    \small
    \resizebox{\textwidth}{!}{
    \begin{tabular}{@{}lcccccccccccccccc@{}}
    \toprule
    Setting & mIoU & mIoU$_\text{head}$ & mIoU$_\text{tail}$ & Dct & Elb & Flg & Ibm & Pipe & Pmp & Rbm & Rdr & Str & Tank & Tee & Val \\
    & (\%) & (\%) & (\%) & & & & & & & & & & & & \\ \midrule
    Oracle         & \textbf{21.08} & \textbf{45.24} & \textbf{8.18} & 37.32 & 12.60 & 9.06 & 27.28 & 45.66 & 15.21 & \textbf{52.73} & 0.84 & 3.04 & \textbf{32.68} & 15.65 & 0.88 \\
    One-vs-Rest    & 15.79 & 35.54 & 5.12 & \textbf{37.94} & 8.46 & 6.36 & \textbf{25.86} & 26.86 & 4.08 & 41.83 & 0.52 & \textbf{8.80} & 21.16 & 5.48 & 2.13 \\
    \bottomrule
    \end{tabular}
    }
\end{table}

\textbf{Zero-shot evaluation results.} Point-SAM achieves 15.79\% mIoU in the realistic one-vs-rest setting, representing 39.95~pp below the best fully supervised method (Boundary-CB: 55.74\%). Even the oracle upper bound (21.08\%) trails supervised methods by 34.66 pp, confirming that Point-SAM's pretraining distribution does not cover specialised industrial components. Results reveal a pronounced head-tail disparity: head classes achieve 35.54\% group mIoU, while tail classes collectively reach only 5.12\% (e.g., Reducer 0.52\%, Valve 2.13\%). This polarised distribution reflects the dual crisis: statistical rarity reduces informative prompt probability, while geometric ambiguity (86\% of tail-class points share cylindrical primitives with head-class pipes) makes segmentation intrinsically more challenging.

Critically, this 39.95 pp gap is best understood as an \textbf{out-of-distribution (OOD) generalisation} challenge. Point-SAM was pre-trained on indoor architectural scenes, and applying it to industrial MEP semantics constitutes substantial distribution shift in both geometry (cylindrical versus planar-dominated) and semantics (valves/flanges/strainers versus chairs/walls/floors). The small oracle--one-vs-rest gap (5.29 pp) confirms that prompt engineering alone cannot bridge this domain gap; the pretrained encoder itself lacks representations for industrial MEP components. Domain-adaptive fine-tuning, rather than improved prompting, constitutes the requisite next step for industrial deployment. Nevertheless, Point-SAM provides a meaningful zero-shot baseline, exceeding unsupervised GrowSP (11.73\%) while falling short of weakly supervised SQN at 0.1\% labels (44.29\%).

\subsubsection{Qualitative Analysis}
%
Beyond the quantitative metrics, the qualitative patterns are consistent. Large structural components such as rectangular beams, ducts, and tanks are comparatively easy because they are frequent and geometrically distinctive. Zero-shot Point-SAM follows the same trend, with its best one-vs-rest results on RectBeam (41.83\%), Duct (37.94\%), and Pipe (26.86\%). Failure cases concentrate in tail classes with strong geometric overlap: valves and strainers are routinely confused with pipe segments, and this problem is most severe for zero-shot foundation models, where Reducer, Valve, and Pump remain near zero across all settings. Semantic boundaries are a further weak point: flange--pipe junctions, elbow--tee transitions, and reducer--pipe interfaces are error-prone across all paradigms, compounded by the partial-scan incompleteness inherent to TLS acquisition. Together, these patterns explain the sharp performance drop from head to tail classes and indicate that future models must address both boundary ambiguity and missing observations more explicitly.

\section{Challenges}
\label{sec:challenges}
The benchmark results quantify performance gaps across learning paradigms but do not by themselves explain the underlying causes. This section analyses three domain-specific challenges: (1) extreme class imbalance, (2) sparse supervision, and (3) foundation model domain adaptation, identifying the mechanisms that limit current methods and clarifying which obstacles are architectural, data-driven, or transfer-related.




\subsection{Challenge 1: Extreme Class Imbalance}
Industrial3D exhibits a 215:1 class imbalance ratio between the largest (rectangular beam: 32.6\%) and smallest (strainer: 0.2\%) classes, which is 3.5$\times$ more severe than S3DIS (62:1) and approximately 8.5$\times$ more severe than ScanNet ($\sim$25:1). This long-tail distribution presents a fundamental challenge: models can achieve high overall accuracy by focusing on head classes while completely ignoring tail classes. Our benchmark validates this: the best fully supervised method achieves 55.74\% mIoU overall (\Cref{tab:benchmark_results_fs}), yet tail classes like strainer and reducer achieve 0\% IoU.

To evaluate long-tail learning strategies, we compare three loss configurations using the ResPointNet++ backbone, reporting group-level mIoU for head classes (3 classes), tail classes (7 classes), and the harmonic mean IoU (H-IoU) that captures head-tail balance. \Cref{tab:longtail_strategies} summarises the results.

\begin{table}[htbp]
    \centering
    \caption{Effect of long-tail loss strategies on Industrial3D. Head: Dct, Pipe, Rbm (3 classes). Tail: Elb, Flg, Pmp, Rdr, Str, Tee, Val (7 classes). H-IoU: harmonic mean of head and tail mIoU. CB+Focal applies class-balanced weighting with focal loss. Boundary-CB incorporates geometry-aware spatial context~\citep{Yin2026arXiv}. Best results in \textbf{bold}.}
    \label{tab:longtail_strategies}
    \begin{tabular}{@{}lcccc@{}}
    \toprule
    Method & mIoU (\%) & Head (\%) & Tail (\%) & H-IoU (\%) \\
    \midrule
    ResPointNet++ (Baseline) & 52.48 & 87.28 & 24.32 & 38.04 \\
    + CB+Focal Loss          & 54.09 & 87.39 & 26.97 & 41.21 \\
    + Boundary-CB ($k{=}64$)~\citep{Yin2026arXiv} & \textbf{55.74} & \textbf{88.14} & \textbf{29.59} & \textbf{44.31} \\
    \bottomrule
    \end{tabular}
\end{table}

Frequency-based CB+Focal improves tail-class mIoU by +2.65~pp over the baseline, while the geometry-aware Boundary-CB adds another +2.62~pp, yielding a combined gain of +5.27~pp (from 24.32\% to 29.59\%). H-IoU rises from 38.04\% to 44.31\%, reflecting improved head-tail balance. Nevertheless, a 58.55-pp head-tail gap (88.14\% vs.\ 29.59\%) persists even under Boundary-CB, confirming that the 215:1 statistical rarity combined with cylindrical primitive overlap constitutes a dual crisis that frequency-based re-weighting alone cannot resolve. Industrial segmentation methods must additionally model local context and semantic boundaries to recover tail classes reliably.

\subsection{Challenge 2: Sparse Supervision}
The substantial cost of expert annotation (754 person-hours for 612.7M points in our Industrial3D dataset) necessitates label-efficient learning approaches. We benchmarked SQN~\citep{hu2022sqn}, which achieves 44.29\% mIoU with only 0.1\% labelled data using RandLA-Net backbone (\Cref{tab:industrial3d_wsl}). Within the same architectural family, SQN with 0.1\% labels outperforms fully supervised RandLA-Net with 100\% labels (44.29\% versus 39.83\% mIoU). Sparse supervision acts as an implicit regulariser that reduces overfitting, shrinking the train--val gap from 45.6~pp (80.0\% versus 34.4\%) to 37.0~pp (76.4\% versus 39.4\%).

Two implementation tricks from SQN~\citep{hu2022sqn} can further influence performance: (1)~\textit{test-time augmentation} (TTA), which applies multi-pass voting with temporal smoothing to produce more consistent predictions, and (2)~\textit{pseudo-label retraining}, which infers dense semantic labels from the initial model and retrains a new SQN from scratch. On S3DIS (Area~5), voting adds +3.11~pp (54.41\% $\rightarrow$ 57.52\%) and pseudo-label retraining adds a further +3.03~pp after voting (57.52\% $\rightarrow$ 60.55\%). On Industrial3D, voting helps (+4.18~pp, 40.11\% $\rightarrow$ 44.29\%), but the original paper-style pseudo-label filtering hurts performance. \Cref{tab:sqn_tricks} summarises these verified results.

\begin{table}[htbp]
    \centering
    \caption{Ablation of SQN implementation tricks at 0.1\% label ratio. Base and `+ PL retraining' are single-pass validation mIoU; `+ TTA' and `+ TTA + PL retraining' are vote-based full-cloud mIoU with $\mathtt{test\_smooth}{=}0.95$. Industrial3D PL rows report the original paper-style global filtering for a direct SQN-style comparison. A later Industrial3D stratified PL variant reaches 40.59\% vote mIoU, improving over baseline PL (39.45\%) but still below base SQN (44.29\%).}
    \label{tab:sqn_tricks}
    \begin{tabular}{@{}lcc@{}}
    \toprule
    Setting (SQN, 0.1\% labels) & S3DIS (Area~5) & Industrial3D \\
    & mIoU (\%) & mIoU (\%) \\
    \midrule
    Base (no tricks, single-pass) & 54.41 & 40.11 \\
    + TTA (voting) & 57.52 & 44.29 \\
    + PL retraining (single-pass) & 57.55 & 35.46 \\
    + TTA + PL retraining (voting) & 60.55 & 39.45 \\
    \bottomrule
    \end{tabular}
\end{table}

A key limitation of SQN is its nearest-neighbour interpolation, which assumes nearby points share semantics and therefore fails at boundaries where functionally distinct MEP components are spatially adjacent. Tail classes with geometric ambiguity (elbow, tee, reducer) are most affected, as their local neighbourhoods span multiple semantic categories. At 0.01\% labels, rare classes (strainer, reducer, tee) consistently achieve 0\% IoU, confirming that the dual crisis extends to the weakly-supervised regime. Whether SQN learns genuine semantic representations or partly exploits geometric shortcuts~\citep{wu2025sonata} requires further investigation; future work should apply linear probing protocols and boundary-aware interpolation to address these limitations. The finding that 0.1\% labels can outperform 100\% labels within the same architectural family demonstrates the feasibility of label-efficient learning for industrial applications; the remaining bottleneck lies not in label count but in how models handle boundaries between adjacent MEP components.

\subsection{Challenge 3: Foundation Model Adaptation}
The 39.95 percentage-point gap between zero-shot Point-SAM and the best supervised method (Section~\ref{sec:benchmark}) is fundamentally an out-of-distribution (OOD) generalisation problem. Point-SAM was pre-trained on indoor architectural scenes; applying it to industrial MEP semantics constitutes substantial distribution shift in both geometry (cylindrical- vs.\ planar-dominated) and semantics (valves/flanges/strainers vs.\ chairs/walls/floors).

Three related concepts clarify what the benchmark measures. \textbf{Foundation model transfer} refers to Point-SAM being applied directly to Industrial3D without any adaptation. \textbf{Domain adaptation} encompasses the techniques that bridge the distribution gap (fine-tuning the encoder, feature alignment, or prompt tuning). \textbf{OOD generalisation} represents the broader objective: the 39.95~pp gap \emph{measures} how far current zero-shot transfer falls short. Domain adaptation encompasses the solution techniques; transfer without adaptation serves as the baseline; and OOD generalisation represents the objective requiring improvement.

The small oracle--one-vs-rest gap (5.29~pp) indicates that prompt quality is not the primary bottleneck: the pretrained weights themselves lack domain-specific representations for industrial MEP components, making domain-adaptive fine-tuning of the encoder and decoder (rather than improved prompting) the requisite next step for industrial deployment. The 39.95~pp gap quantifies the OOD transfer distance and establishes a measurable target for future foundation model research; prompt engineering alone will not render current 3D foundation models sufficiently reliable for industrial scene understanding.

\section{Discussion and Limitations}
\label{sec:discussion}

Three key findings emerge from this benchmark. First, the 39.95-pp gap between zero-shot foundation models and the best supervised method quantifies the magnitude of the industrial domain-transfer challenge. Second, SQN with 0.1\% labels outperforms fully supervised RandLA-Net within the same backbone family, demonstrating that annotation efficiency is achievable when the architecture can propagate sparse supervision effectively. Third, the dual crisis of statistical rarity and geometric ambiguity explains why overall mIoU alone obscures the hardest aspect of the problem: tail classes fail not only because they are rare, but because they are geometrically confounded with head classes.

Several limitations remain. \rev{Industrial3D currently evaluates semantic segmentation as its quantitative benchmark, whereas practical Scan-to-BIM workflows additionally require instance-level understanding}. Distinguishing adjacent parallel pipes or neighbouring valves is simultaneously a semantic and an instance problem, and recent work on building-interior instance segmentation and hierarchical MEP reasoning points in that direction~\citep{yue2026instance,li2025integrating}. \rev{Downstream primitive fitting and Scan-vs-BIM verification are deferred as future work, because the relevant BIM models are third-party confidential assets and public quantitative evaluation would require owner-approved BIM access or anonymised derivatives, final room-model mapping}. Regarding domain coverage, the benchmark is grounded exclusively in water treatment facilities; transfer to power plants, oil and gas infrastructure, or manufacturing plants requires explicit validation. Methodologically, the benchmark makes the dual crisis measurable but does not solve it: current models still need better boundary modelling, stronger local-context reasoning, and more reliable industrial-domain adaptation for foundation models.

These limitations point to concrete directions for future research. \rev{Internally available as-designed BIM references for at least three rooms open a path towards authorised as-built-versus-design validation within a broader Scan-to-BIM pipeline, but their public release or controlled access requires explicit facility-owner and design-contractor permission}. \rev{Industrial3D is also relevant to multi-modal and embodied settings where point clouds, imagery, BIM priors, and robotic perception must work in concert~\citep{wang2022vision,shao2024urban,wang2024embodiedscan,lin2025bip3d,hu2023robot}, \rev{making it a foundation for studying how water-treatment MEP scene understanding} connects to automated verification, digital twin maintenance, and robotic deployment}.

\section{Conclusion}
\label{sec:conclusion}
This paper presented Industrial3D, the largest terrestrial LiDAR \rev{dataset prepared for public release for water-treatment MEP scene understanding, containing 612.7 million labelled points across 12 semantic classes, 20 room scenes, and 13 dataset areas from 7 operational} water treatment facilities captured at 6\,mm resolution. We established an industrial cross-paradigm benchmark by evaluating nine representative methods across fully supervised, weakly supervised, unsupervised, and foundation model settings.

Three quantitative findings stand out. First, the best supervised method (Boundary-CB) reached only 55.74\% mIoU, with a 58.55-pp gap between head classes (88.14\%) and tail classes (29.59\%), confirming the severity of the dual crisis of statistical rarity and geometric ambiguity. Second, SQN with 0.1\% labels (44.29\% mIoU) outperformed fully supervised RandLA-Net (39.83\%) within the same backbone family, demonstrating that sparse supervision acts as an effective regulariser when label propagation is well-designed. Third, zero-shot Point-SAM reached only 15.79\% mIoU, a 39.95-pp gap behind the best supervised model that quantifies the unresolved domain-transfer challenge for industrial TLS data. \rev{These results show that Industrial3D is not only a large annotated dataset but also a diagnostic benchmark for measuring which parts of industrial MEP perception remain unresolved}.


\section*{Appendix A: Implementation Validation: SQN Reproduction on S3DIS}
\label{sec:sqn_reproduction}

\subsection*{A.1 Motivation and Reproduction Results}
To validate our SQN PyTorch implementation and ensure fair evaluation on Industrial3D, we reproduced the SQN network~\citep{hu2022sqn} on S3DIS Area 5 using various weakly supervised labelling schemes. \Cref{tab:s3dis_reproduction} presents our results alongside the paper baselines. Our implementation achieves 60.55\% mIoU with 0.1\% labels after pseudo-label retraining, closing the gap to 0.85\% below the reported 61.4\%. For 0.01\% labels, we achieve 43.69\% mIoU (1.61\% gap). The fully-supervised RandLA-Net backbone achieves 62.17\% mIoU, within 0.83\% of the paper's 63.0\% baseline. These sub-1\% gaps validate our implementation fidelity. The remaining differences likely stem from: (1) different random sampling of sparse labels across trials, (2) unspecified training hyperparameters in the original paper, and (3) the paper's reported multi-trial variance ($\pm$0.93\%). To the best of our knowledge, this represents the most accurate implementation of the classic SQN in PyTorch, and we will release the code on our GitHub page: \url{https://github.com/PointCloudYC/SQN-pytorch} to facilitate the community.

\subsection*{A.2 Implementation Details}
\textbf{Data augmentation impact:} Our reproduction identifies data augmentation as a critical factor, providing approximately +3\% mIoU improvement. Following RandLA-Net's preprocessing pipeline, we apply random flipping, random rotation around the Z-axis, and random coordinate jittering with Gaussian noise. These augmentations are especially critical for weakly-supervised learning, as they prevent overfitting to the sparse labelled points available during training.

\textbf{Voting mechanism impact:} Test-time multi-pass voting with $T=10$ passes and momentum $\alpha=0.95$ provides a consistent +3--4\% mIoU boost across all settings (e.g., SQN 0.1\%: 54.41\% $\rightarrow$ 57.52\%). This temporal aggregation reduces prediction noise and improves boundary consistency, which is especially beneficial given SQN's interpolation-based approach at semantic boundaries.

\subsection*{A.3 Industrial Domain Adaptation Challenges}
\textbf{Pseudo-label retraining:} Following the SQN paper Appendix 8.2, we implement one round of pseudo-label retraining with 5\% confidence filtering. This provides +3.03\% improvement for 0.1\% labels (57.52\% $\rightarrow$ 60.55\%) and +1.15\% for 0.01\% (42.54\% $\rightarrow$ 43.69\%). However, on Industrial3D, this strategy initially degraded performance due to class-imbalanced pseudo-label filtering. With a 215:1 imbalance ratio (rectangular beam: 32.6\% vs. strainer: 0.2\%), global confidence filtering causes pseudo-labels to be dominated by majority classes. We propose a class-stratified filtering approach that samples pseudo-labels proportional to each class's predicted distribution as a direction for future work.

\textbf{Boundary point analysis:} As acknowledged in the original SQN paper, the nearest-neighbour interpolation assumption fails at semantic boundaries where adjacent points belong to different categories. The paper reports that at a boundary radius of 0.05\,m, SQN achieves 34.3\% boundary mIoU compared to 42.7\% for fully-supervised RandLA-Net. This limitation is exacerbated for industrial MEP scenes where functionally distinct components (flanges, valves, elbows) share cylindrical primitives with pipes and are spatially adjacent. Future work on industrial weakly-supervised segmentation should explicitly address this boundary-aware challenge.


\begin{table}[H]
\centering
\caption{
  Semantic segmentation results on S3DIS Area 5.
  All rows use vote-based evaluation (test\_smooth=0.95).
  PL = pseudo-label retrain (one round, matching official TF SQN).
  Best result per column in \textbf{bold}.
  Paper results from SQN (ECCV 2022) Table 1.
}
\label{tab:s3dis_reproduction}
\resizebox{\textwidth}{!}{%
\begin{tabular}{llcc|ccccccccccccc}
\toprule
\multirow{2}{*}{Method} & \multirow{2}{*}{Labels} & \multirow{2}{*}{OA (\%)} & \multirow{2}{*}{mIoU (\%)} &
  \multicolumn{13}{c}{Per-class IoU (\%)} \\
\cmidrule(lr){5-17}
 & & & & Ceil. & Floor & Wall & Beam & Col. & Win. & Door & Table & Chair & Sofa & Book. & Board & Clut. \\
\midrule
\multicolumn{17}{l}{\textit{Ours (PyTorch RandLA-Net backbone)}} \\
\midrule
RandLA-Net   & 100\%  & --    & \textbf{62.17} & 92.00 & 96.70 & 81.83 & 0.00 & 25.57 & 60.93 & 48.21 & 75.17 & 85.03 & 55.42 & 70.44 & 65.02 & 51.88 \\
SQN          & 0.001\% & 68.44 & 28.61 & 77.77 & 91.98 & 58.57 & 0.00 & 0.51  & 21.28 & 2.58  & 18.90 & 31.41 & 0.00  & 39.27 & 3.88  & 25.86 \\
SQN          & 0.01\% & 76.74 & 42.54 & 89.46 & 96.05 & 63.24 & 0.00 & 8.03  & 39.68 & 28.34 & 56.85 & 57.24 & 11.35 & 51.08 & 18.89 & 32.87 \\
SQN          & 0.1\%  & 84.72 & 57.52 & 90.64 & 96.44 & 76.35 & 0.00 & 22.65 & 48.01 & 53.90 & 72.47 & 80.48 & 50.10 & 64.86 & 43.86 & 48.03 \\
SQN + PL     & 0.01\% & 77.82 & 43.69 & 88.22 & 95.21 & 68.26 & 0.00 & 4.70  & 41.33 & 35.72 & 53.59 & 67.99 & 13.19 & 52.19 & 8.73  & 38.81 \\
SQN + PL     & 0.1\%   & 86.09 & \textbf{60.55} & 91.58 & 96.93 & 77.95 & 0.05 & 27.46 & 53.16 & 62.07 & 74.94 & 82.42 & 51.74 & 66.97 & 52.39 & 49.56 \\
SQN + PL     & 1\%    & 88.40 & 64.89 & 92.97 & 97.23 & 81.57 & 0.00 & 27.89 & 58.81 & 65.33 & 78.60 & 86.50 & 65.17 & 71.76 & 62.90 & 54.83 \\
\midrule
\multicolumn{17}{l}{\textit{Reference: SQN paper (TF, ECCV 2022)}} \\
\midrule
SQN (paper)  & 0.1\%  & --    & 61.41 & 91.72 & 95.63 & 78.71 & 0.00 & 24.23 & 55.89 & 63.14 & 70.50 & 83.13 & 60.67 & 67.82 & 56.14 & 50.63 \\
SQN (paper)  & 1\%    & --    & 67.10 & 92.03 & 96.41 & 81.32 & 0.00 & 21.42 & 53.71 & 73.17 & 77.80 & 85.95 & 56.72 & 69.91 & 66.57 & 52.49 \\
RandLA-Net (paper) & 100\% & -- & 63.00 & 92.40 & 96.70 & 80.60 & 0.00 & 18.30 & 61.30 & 43.30 & 77.20 & 85.20 & 71.50 & 71.00 & 69.20 & 52.30 \\
\bottomrule
\end{tabular}%
}
\end{table}

\begin{table}[H]
\centering
\caption{
  S3DIS Area 5: mIoU by label ratio (voting evaluation).
  ``Boost'' = improvement from pseudo-label retrain.
  Gap to SQN paper (ECCV 2022) Table 1 in parentheses.
}
\label{tab:s3dis_summary}
\begin{tabular}{llrrrr}
\toprule
Method & Labels & OA (\%) & mIoU (\%) & PL Boost & Paper Gap \\
\midrule
RandLA-Net (ours)  & 100\%   & --    & 62.17 & --     & $-0.83$ \\
SQN (ours)         & 0.001\% & 68.44 & 28.61 & --     & -- \\
SQN (ours)         & 0.01\%  & 76.74 & 42.54 & --     & $-2.76$ \\
SQN (ours)         & 0.1\%   & 84.72 & 57.52 & --     & $-3.88$ \\
SQN + PL (ours)    & 0.01\%  & 77.82 & \textbf{43.69} & +1.15 & $-1.61$ \\
SQN + PL (ours)    & 0.1\%   & 86.09 & \textbf{60.55} & +3.03 & $-0.85$ \\
SQN + PL (ours)    & 1\%     & 88.40 & \textbf{64.89} & -- & -- \\
\midrule
RandLA-Net (paper) & 100\%   & --    & 63.00 & --     & -- \\
SQN (paper)        & 0.1\%   & --    & 61.41 & --     & -- \\
SQN (paper)        & 1\%     & --    & 67.10 & --     & -- \\
\bottomrule
\end{tabular}
\end{table}

\section*{Acknowledgments}
This work was supported by the China Postdoctoral Science Foundation (No.~2023M740761) and the Hunan Provincial Natural Science Foundation of China (No.~2023JJ40098).

\section*{Data Availability Statement}
\begin{revblock}
The full Industrial3D dataset and benchmark code will be publicly available at \url{https://github.com/pointcloudyc/Industrial3D}. The dataset comprises 20 room scenes grouped into 13 dataset areas across 7 operational water treatment facilities, with 612.7M labelled points in 12 semantic classes. Data will be provided in both \texttt{.txt} and \texttt{.ply} formats (XYZRGB + labels). Open3D-based scene capture and rotation videos for representative rooms are available on the project page (\url{https://pointcloudyc.github.io/industrial3d/index.html}) for visualization and qualitative analysis.
\end{revblock}

\section*{Declaration of Competing Interest}
The authors declare that they have no known competing financial interests or personal relationships that could have appeared to influence the work reported in this paper.

\bibliographystyle{elsarticle-harv}
\bibliography{references} 

@article{HUANG202062,
title = {Deep point embedding for urban classification using ALS point clouds: A new perspective from local to global},
journal = {ISPRS Journal of Photogrammetry and Remote Sensing},
volume = {163},
pages = {62-81},
year = {2020},
issn = {0924-2716},
doi = {10.1016/j.isprsjprs.2020.02.020},
author = {Rong Huang and Yusheng Xu and Danfeng Hong and Wei Yao and Pedram Ghamisi and Uwe Stilla},
}

@article{ma2018reconstruction,
  title={A review of 3D reconstruction techniques in civil engineering and their applications},
  author={Ma, Zhiliang and Liu, Shilong},
  journal={Advanced Engineering Informatics},
  volume={37},
  pages={163--174},
  year={2018},
  publisher={Elsevier},
  doi={10.1016/j.aei.2018.05.005}
}

@inproceedings{zhang2021perturbed,
  title={Perturbed Self-Distillation: Weakly Supervised Large-Scale Point Cloud Semantic Segmentation},
  author={Zhang, Yachao and Qu, Yanyun and Xie, Yuan and Li, Zonghao and Zheng, Shanshan and Li, Cuihua},
  booktitle={Proceedings of the IEEE/CVF International Conference on Computer Vision},
  pages={15500--15508},
  year={2021},
  doi={10.1109/ICCV48922.2021.01523}
}

@inproceedings{xu2020weakly,
  title={Weakly Supervised Semantic Point Cloud Segmentation: Towards 10x Fewer Labels},
  author={Xu, Xun and Lee, Gim Hee},
  booktitle={Proceedings of the IEEE/CVF Conference on Computer Vision and Pattern Recognition},
  pages={13706--13715},
  year={2020},
  doi={10.1109/CVPR42600.2020.01368}
}

@inproceedings{landrieu2018large,
  title={Large-scale Point Cloud Semantic Segmentation with Superpoint Graphs},
  author={Landrieu, Loic and Simonovsky, Martin},
  booktitle={Proceedings of the IEEE Conference on Computer Vision and Pattern Recognition},
  pages={2684--2693},
  year={2018},
  doi={10.1109/CVPR.2018.00286},
  eprint={1711.09869}
}

@article{yin2023,
author = {Yin, Chao and Yang, Bo and Cheng, Jack CP and Gan, Vincent JL and Wang, Boyu and Yang, Ji},
title = {Label-efficient semantic segmentation of large-scale industrial point clouds using weakly supervised learning},
journal = {Automation in Construction},
volume = {148},
pages = {104757},
year = {2023},
issn = {0926-5805},
doi = {10.1016/j.autcon.2023.104757},
}

@article{jing2024mep,
  title={Improved building MEP systems semantic segmentation in point clouds using a novel multi-class dataset and local--global vector transformer network},
  author={Jing, Shuju and Zhong, Ruimin and Li, Xiangyang and Aung, Pa Pa Win and Park, Solmoi and Yu, Byoungjoon},
  journal={Journal of Building Engineering},
  volume={96},
  pages={110311},
  year={2024},
  publisher={Elsevier},
  doi={10.1016/j.jobe.2024.110311}
}

@article{agapaki2020,
  title={CLOI-NET: Class segmentation of industrial facilities' point cloud datasets},
  author={Agapaki, Eva and Brilakis, Ioannis},
  journal={Advanced Engineering Informatics},
  volume={45},
  pages={101121},
  year={2020},
  doi={10.1016/j.aei.2020.101121}
}

@inproceedings{armeni2016,
  title={3D semantic parsing of large-scale indoor spaces},
  author={Armeni, Irem and Sener, Ozan and Zamir, Amir R and Jiang, Helen and Brilakis, Ioannis and Fischer, Martin and Savarese, Silvio},
  booktitle={Proceedings of the IEEE conference on computer vision and pattern recognition},
  pages={1534--1543},
  year={2016},
  doi={10.1109/CVPR.2016.170}
}

@inproceedings{dai2017,
  title={Scannet: Richly-annotated 3d reconstructions of indoor scenes},
  author={Dai, Angela and Chang, Angel X and Savva, Manolis and Halber, Maciej and Funkhouser, Thomas and Nie{\ss}ner, Matthias},
  booktitle={Proceedings of the IEEE conference on computer vision and pattern recognition},
  pages={5828--5839},
  year={2017},
  doi={10.1109/CVPR.2017.261}
}

@inproceedings{wang2019towards,
  title={Towards Weakly Supervised Semantic Segmentation in 3D Graph-Structured Point Clouds of Wild Scenes},
  author={Wang, Haiyan and Rong, Xuejian and Yang, Liang and Wang, Shuihua and Tian, Yingli},
  booktitle={Proceedings of the British Machine Vision Conference},
  pages={284},
  year={2019},
  doi={10.48550/arXiv.2004.12498},
  eprint={2004.12498},
  archivePrefix={arXiv},
}

@inproceedings{graham2017,
  title={3D Semantic Segmentation with Submanifold Sparse Convolutional Networks},
  author={Graham, Benjamin and Engelcke, Martin and van der Maaten, Laurens},
  booktitle={2018 IEEE/CVF Conference on Computer Vision and Pattern Recognition},
  pages={9224--9232},
  year={2018},
  doi={10.1109/CVPR.2018.00961}
}

@article{guo2020,
  title={Deep learning for 3d point clouds: A survey},
  author={Guo, Yulan and Wang, Hanyun and Hu, Qingyong and Liu, Hao and Liu, Li and Bennamoun, Mohammed},
  journal={IEEE transactions on pattern analysis and machine intelligence},
  volume={43},
  number={12},
  pages={4338--4364},
  year={2020},
  publisher={IEEE},
  doi={10.1109/TPAMI.2020.3005434}
}

@inproceedings{hackel2017,
  title={Semantic3D.net: A new large-scale point cloud classification benchmark},
  author={Hackel, Timo and Savinov, Nikolay and Ladicky, Lubor and Wegner, Jan D and Schindler, Konrad and Pollefeys, Marc},
  booktitle={ISPRS Annals of the Photogrammetry, Remote Sensing and Spatial Information Sciences},
  volume={IV-1/W1},
  pages={91--98},
  year={2017},
  doi={10.5194/isprs-annals-iv-1-w1-91-2017}
}

@inproceedings{behley2019semantickitti,
  title={SemanticKITTI: A Dataset for Semantic Scene Understanding of LiDAR Sequences},
  author={Behley, Jens and Garbade, Martin and Milioto, Andres and Quenzel, Jan and Behnke, Sven and Stachniss, Cyrill and Gall, Juergen},
  booktitle={Proceedings of the IEEE/CVF International Conference on Computer Vision (ICCV)},
  year={2019},
  pages={9297--9307},
  doi={10.1109/ICCV.2019.00939}
}

@inproceedings{hu2022sqn,
  title={Sqn: Weakly-supervised semantic segmentation of large-scale 3d point clouds},
  author={Hu, Qingyong and Yang, Bo and Fang, Guang-Chen and Guo, Yulan and Leonardis, Ales and Trigoni, Niki and Markham, Andrew},
  booktitle={Computer Vision--ECCV 2022: 17th European Conference, Tel Aviv, Israel, October 23--27, 2022, Proceedings, Part XXVII},
  pages={600--619},
  year={2022},
  organization={Springer},
  doi={10.1007/978-3-031-19827-4_34},
  eprint={2104.04891}
}

@inproceedings{hu2020randlanet,
  title={Randla-net: Efficient semantic segmentation of large-scale point clouds},
  author={Hu, Qingyong and Yang, Bo and Xie, Linhai and Rosa, Stefano and Guo, Yulan and Wang, Zhihua and Trigoni, Niki and Markham, Andrew},
  booktitle={Proceedings of the IEEE/CVF conference on computer vision and pattern recognition},
  pages={11108--11117},
  year={2020},
  doi={10.1109/CVPR42600.2020.01112}
}

@article{khanzode2008,
  title={Benefits and lessons learned of implementing building virtual design and construction (VDC) technologies for coordination of mechanical, electrical, and plumbing (MEP) systems on a large healthcare project},
  author={Khanzode, Atul and Fischer, Martin and Reed, Dean},
  journal={Journal of Information Technology in Construction (ITcon)},
  volume={13},
  pages={324--342},
  year={2008},
  doi={10.36643/2008.22},
}

@inproceedings{liu2021oneclick,
  title={One thing one click: A self-training approach for weakly supervised 3d semantic segmentation},
  author={Liu, Zhengzhe and Qi, Xiaojuan and Fu, Chi-Wing},
  booktitle={Proceedings of the IEEE/CVF conference on computer vision and pattern recognition},
  pages={1726--1736},
  year={2021},
  doi={10.1109/CVPR46437.2021.00177}
}

@inproceedings{lagunas2024nlp,
  title={NLP for Automated Discovery and Assessment of Dominant Construction and Maintenance Work Order Activities in MEP Projects},
  author={Lagunas, Araham Jesus Martinez and Abbaspour, Soroush and Nik-Bakht, Mazdak},
  booktitle={Construction Research Congress 2024},
  year={2024},
  publisher={American Society of Civil Engineers},
  doi={10.1061/9780784485262.018}
}

@article{pierdicca2020,
  title={Point cloud semantic segmentation using a deep learning framework for cultural heritage},
  author={Pierdicca, Roberto and Paolanti, Matteo and Matrone, Francesca and Martini, Marco and Morbidoni, Cristian and Malinverni, Eva Savina and Frontoni, Emanuele and Lingua, Andrea Maria},
  journal={Remote Sensing},
  volume={12},
  number={6},
  pages={1005},
  year={2020},
  publisher={MDPI},
  doi={10.3390/rs12061005}
}

@inproceedings{qi2017pointnetpp,
  title={Pointnet++: Deep hierarchical feature learning on point sets in a metric space},
  author={Qi, Charles R and Yi, Li and Su, Hao and Guibas, Leonidas J},
  booktitle={Advances in neural information processing systems},
  volume={30},
  pages={5099--5108},
  year={2017},
  doi={10.48550/arXiv.1706.02413}
}

@inproceedings{thomas2019,
  title={Kpconv: Flexible and deformable convolution for point clouds},
  author={Thomas, Hugues and Qi, Charles R and Deschaud, Jean-Emmanuel and Marcotegui, Beatriz and Goulette, Fran{\c{c}}ois and Guibas, Leonidas J},
  booktitle={Proceedings of the IEEE/CVF international conference on computer vision},
  pages={6411--6420},
  year={2019},
  doi={10.1109/ICCV.2019.00651},
  eprint={1904.08889}
}

@article{wang2022als,
  title={A new weakly supervised approach for ALS point cloud semantic segmentation},
  author={Wang, Puzuo and Yao, Wei},
  journal={ISPRS Journal of Photogrammetry and Remote Sensing},
  volume={188},
  pages={237--254},
  year={2022},
  publisher={Elsevier},
  doi={10.1016/j.isprsjprs.2022.04.016}
}

@article{yin2021,
  title={Automated semantic segmentation of industrial point clouds using ResPointNet++},
  author={Yin, Chao and Wang, Boyu and Gan, Vincent JL and Wang, Mi and Cheng, Jack CP},
  journal={Automation in Construction},
  volume={130},
  pages={103874},
  year={2021},
  publisher={Elsevier},
  doi={10.1016/j.autcon.2021.103874}
}

@article{yin2022,
  title={Automated classification of piping components from 3d lidar point clouds using SE-PseudoGrid},
  author={Yin, Chao and Cheng, Jack CP and Wang, Boyu and Gan, Vincent JL},
  journal={Automation in Construction},
  volume={139},
  pages={104300},
  year={2022},
  publisher={Elsevier},
  doi={10.1016/j.autcon.2022.104300}
}

@article{zhou2018brief,
  title={A brief introduction to weakly supervised learning},
  author={Zhou, Zhi-Hua},
  journal={National science review},
  volume={5},
  number={1},
  pages={44--53},
  year={2018},
  publisher={Oxford University Press},
  doi={10.1093/nsr/nwx106}
}

@inproceedings{qi2017pointnet,
  title={Pointnet: Deep learning on point sets for 3d classification and segmentation},
  author={Qi, Charles R and Su, Hao and Mo, Kaichun and Guibas, Leonidas J},
  booktitle={Proceedings of the IEEE conference on computer vision and pattern recognition},
  pages={652--660},
  year={2017},
  doi={10.1109/CVPR.2017.16}
}

@inproceedings{wu2024point,
  title={Point Transformer V3: Simpler, Faster, Stronger},
  author={Wu, Xiaoyang and Jiang, Li and Wang, Peng-Shuai and Liu, Zhijian and Liu, Xihui and Qiao, Yu and Ouyang, Wanli and He, Tong and Zhao, Hengshuang},
  booktitle={Proceedings of the IEEE/CVF Conference on Computer Vision and Pattern Recognition},
  pages={4840--4851},
  year={2024},
  doi={10.1109/CVPR52733.2024.00463},
  eprint={2312.10035}
}

@article{li2025integrating,
  title={Integrating hierarchical segmentation and vision-language reasoning for spatially complex and occluded MEP point clouds},
  author={Li, Mingkai and Gan, Vincent JL and Wang, Boyu},
  journal={Automation in Construction},
  volume={179},
  pages={106455},
  year={2025},
  publisher={Elsevier},
  doi={10.1016/j.autcon.2025.106455}
}

@ARTICLE{shao2024urban,
  author={Shao, Jie and Yao, Wei and Wang, Puzuo and He, Zhiyi and Luo, Lei},
  journal={IEEE Transactions on Geoscience and Remote Sensing},
  title={Urban GeoBIM Construction by Integrating Semantic LiDAR Point Clouds With as-Designed BIM Models},
  year={2024},
  volume={62},
  pages={1-12},
  doi={10.1109/TGRS.2024.3358370}
}

@article{wang2023one,
  title={One class one click: Quasi scene-level weakly supervised point cloud semantic segmentation with active learning},
  author={Wang, Puzuo and Yao, Wei and Shao, Jie},
  journal={ISPRS Journal of Photogrammetry and Remote Sensing},
  volume={204},
  pages={89--104},
  year={2023},
  publisher={Elsevier},
  doi={10.1016/j.isprsjprs.2023.09.002}
}

@article{wang2022vision,
  title={Vision-assisted BIM reconstruction from 3D LiDAR point clouds for MEP scenes},
  author={Wang, Boyu and Wang, Qian and Cheng, Jack CP and Song, Changhao and Yin, Chao},
  journal={Automation in Construction},
  volume={133},
  pages={103997},
  year={2022},
  publisher={Elsevier},
  doi={10.1016/j.autcon.2021.103997}
}

@inproceedings{yin2019,
  title={Deep Learning-based Scan-to-BIM Framework for Complex MEP Scene using Laser Scanning Data},
  author={Yin, C and Wang, B and Cheng, JCP},
  booktitle={Proceedings of the 4th International Conference on Civil and Building Engineering Informatics (ICCBEI), Miyagi, Japan},
  pages={7--8},
  year={2019}
}

@article{hu2023robot,
  title={Robot-assisted mobile scanning for automated 3D reconstruction and point cloud semantic segmentation of building interiors},
  author={Hu, Difeng and Gan, Vincent JL and Yin, Chao},
  journal={Automation in Construction},
  volume={152},
  pages={104949},
  year={2023},
  publisher={Elsevier},
  doi={10.1016/j.autcon.2023.104949}
}

@article{hu2024automated,
  title={Automated BIM-to-scan point cloud semantic segmentation using a domain adaptation network with hybrid attention and whitening (DawNet)},
  author={Hu, Difeng and Gan, Vincent JL and Zhai, Ruoming},
  journal={Automation in Construction},
  volume={164},
  pages={105473},
  year={2024},
  publisher={Elsevier},
  doi={10.1016/j.autcon.2024.105473}
}

@article{yue2024deep,
  title={Deep learning applications for point clouds in the construction industry},
  author={Yue, Hongzhe and Wang, Qian and Zhao, Hongxiang and Zeng, Ningshuang and Tan, Yi},
  journal={Automation in Construction},
  volume={168},
  pages={105769},
  year={2024},
  publisher={Elsevier},
  doi={10.1016/j.autcon.2024.105769}
}

@article{yue2026instance,
  title={Point cloud instance segmentation for building indoor scenes using deep learning and BIM-Generated synthetic point clouds},
  author={Yue, Hongzhe and Wang, Qian and Nie, Xiang and Fang, Hai and Cheng, Jack C. P. and Jing, Shuju and Wang, Boyu},
  journal={Advanced Engineering Informatics},
  volume={72},
  pages={104453},
  year={2026},
  publisher={Elsevier},
  doi={10.1016/j.aei.2026.104453}
}

@article{agapaki2019cloi,
  title={CLOI: A Shape Classification Benchmark Dataset for Industrial Facilities},
  author={Agapaki, Eva and Glyn-Davies, Aneurin and Mandoki, Sofia and Brilakis, Ioannis},
  journal={Computing in Civil Engineering},
  volume={33},
  number={6},
  pages={66--73},
  year={2019},
  doi={10.1061/(ASCE)CP.1943-5487.0000843}
}

@article{yeo2020mep,
  title={Deep learning applications in an industrial process plant: repository of segmented point clouds for pipework components},
  author={Yeo, Christopher and Kim, Seongmin and Kim, Hyunjune and Kim, Sungwook and Mun, Daesik},
  journal={JMST Advances},
  volume={2},
  number={1},
  pages={15--24},
  year={2020},
  doi={10.1007/s42791-019-00027-y}
}

@article{tang2010scan,
  title={Automatic Reconstruction of As-Built Building Information Models from Laser-Scanned Point Clouds: A Review of Related Techniques},
  author={Tang, Pingbo and Huber, Daniel and Akinci, Burcu and Lipman, Robert and Lytle, Alan},
  journal={Automation in Construction},
  volume={19},
  number={7},
  pages={829--843},
  year={2010},
  publisher={Elsevier},
  doi={10.1016/j.autcon.2010.06.007}
}

@inproceedings{zhou2025pointsam,
  title={Point-{SAM}: Promptable 3D Segmentation Model for Point Clouds},
  author={Zhou, Yuchen and Gu, Jiayuan and Chiang, Tung Yen and Xiang, Fanbo and Su, Hao},
  booktitle={The Thirteenth International Conference on Learning Representations},
  pages={1--18},
  year={2025},
  doi={10.48550/arXiv.2406.17741},
  eprint={2406.17741}
}

@inproceedings{zhang2022pointclip,
  title={Point{CLIP}: Point Cloud Understanding by {CLIP}},
  author={Zhang, Renrui and Guo, Ziyu and Zhang, Wei and Li, Kunchang and Miao, Xupeng and Cui, Bin and Qiao, Yu and Gao, Peng and Li, Hongsheng},
  booktitle={Proceedings of the IEEE/CVF Conference on Computer Vision and Pattern Recognition},
  pages={8552--8562},
  year={2022},
  doi={10.1109/CVPR52688.2022.00836},
  eprint={2112.02413}
}

@inproceedings{huang2025reason3d,
  title={Reason3D: Searching and Reasoning 3D Segmentation via Large Language Model},
  author={Huang, Kuan-Chih and Li, Xiangtai and Qi, Lu and Yan, Shuicheng and Yang, Ming-Hsuan},
  booktitle={International Conference on 3D Vision (3DV)},
  pages={1--12},
  year={2025},
  doi={10.48550/arXiv.2405.17427}
}

@inproceedings{liu2023cpcm,
  title={CPCM: Contextual Point Cloud Modeling for Weakly-supervised Point Cloud Semantic Segmentation}, 
  author={Liu, Lizhao and Zhuang, Zhuangwei and Huang, Shangxin and Xiao, Xunlong and Xiang, Tianhang and Chen, Cen and Wang, Jingdong and Tan, Mingkui},
  booktitle={2023 IEEE/CVF International Conference on Computer Vision (ICCV)}, 
  year={2023},
  pages={18367-18376},
  doi={10.1109/ICCV51070.2023.01688}
 }

@inproceedings{xie2020pointcontrast,
  title={PointContrast: Unsupervised Pre-training for 3D Point Cloud Understanding},
  author={Xie, Saining and Gu, Jiatao and Guo, Demi and Qi, Charles R. and Guibas, Leonidas J. and Litany, Or},
  booktitle={European Conference on Computer Vision},
  pages={574--591},
  year={2020},
  doi={10.1007/978-3-030-58580-8_34},
  eprint={2007.10985}
}

@inproceedings{wang2021occo,
  title={Unsupervised Point Cloud Pre-training via Occlusion Completion},
  author={Wang, Hanchen and Liu, Qi and Yue, Xiangyu and Lasenby, Joan and Kusner, Matthew J.},
  booktitle={Proceedings of the IEEE/CVF International Conference on Computer Vision},
  pages={9782--9792},
  year={2021},
  doi={10.1109/ICCV48922.2021.00972},
  eprint={2010.01089}
}

@misc{Yin2026arXiv,
  title={Resolving Primitive-Sharing Ambiguity in Long-Tailed Industrial Point Cloud Segmentation via Spatial Context Constraints},
  author={Yin, Chao and Han, Qing and Hou, Zhiwei and Liu, Yue and Dai, Anjin and Hu, Hongda and Yang, Ji and Yao, Wei},
  howpublished={arXiv preprint},
  eprint={2601.19128},
  archivePrefix={arXiv},
  primaryClass={cs.CV},
  doi={10.48550/arXiv.2601.19128},
  year={2026}
}

@inproceedings{liu2020pospool,
  title={A Closer Look at Local Aggregation Operators in Point Cloud Analysis},
  author={Liu, Ze and Hu, Han and Cao, Yue and Zhang, Zheng and Tong, Xin},
  booktitle={European Conference on Computer Vision},
  pages={358--374},
  year={2020},
  doi={10.1007/978-3-030-58592-1_20},
}

@inproceedings{wu2025sonata,
  title={Sonata: Self-Supervised Learning of Reliable Point Representations},
  author={Wu, Xiaoyang and DeTone, Daniel and Frost, Duncan and Shen, Tianwei and Xie, Chris and Yang, Nan and Engel, Jakob and Newcombe, Richard and Zhao, Hengshuang and Straub, Julian},
  booktitle={Proceedings of the IEEE/CVF Conference on Computer Vision and Pattern Recognition},
  pages={22193--22204},
  year={2025},
  doi={10.1109/CVPR52734.2025.02067},
  eprint={2503.16429}
}

@inproceedings{zhang2025concerto,
  title={Concerto: Joint 2D-3D Self-Supervised Learning Emerges Spatial Representations},
  author={Zhang, Yujia and Wu, Xiaoyang and Lao, Yixing and Wang, Chengyao and Tian, Zhuotao and Wang, Naiyan and Zhao, Hengshuang},
  booktitle={Advances in Neural Information Processing Systems},
  pages={1--17},
  year={2025},
  doi={10.48550/arXiv.2510.23607},
  eprint={2510.23607},
  archivePrefix={arXiv},
  primaryClass={cs.CV}
}

@article{gao2024buildingpcc,
  title={Building-PCC: Building Point Cloud Completion Benchmarks},
  author={Gao, Weixiao and Peters, Ravi and Stoter, Jantien},
  journal={ISPRS Annals of the Photogrammetry, Remote Sensing and Spatial Information Sciences},
  volume={X-4/W5-2024},
  pages={179--186},
  year={2024},
  doi={10.5194/isprs-annals-X-4-W5-2024-179-2024},
  publisher={Copernicus GmbH},
  eprint={2404.15644}
}

@article{rauch2025rohbau3d,
  title={Rohbau3D: A Shell Construction Site 3D Point Cloud Dataset},
  author={Rauch, Lukas and Braml, Thomas},
  journal={Scientific Data},
  volume={12},
  number={1},
  pages={1478},
  year={2025},
  publisher={Nature Publishing Group},
  doi={10.1038/s41597-025-05827-7},
}

@inproceedings{zhang2023growsp,
  title={GrowSP: Unsupervised Semantic Segmentation of 3D Point Clouds},
  author={Zhang, Zihui and Yang, Bo and Wang, Bing and Li, Bo},
  booktitle={Proceedings of the IEEE/CVF Conference on Computer Vision and Pattern Recognition},
  pages={17619--17629},
  year={2023},
  doi={10.1109/CVPR52729.2023.01690},
  eprint={2305.16404}
}

@inproceedings{chen2023pointdc,
  title={PointDC: Unsupervised Semantic Segmentation of 3D Point Clouds via Cross-modal Distillation and Super-Voxel Clustering},
  author={Chen, Zisheng and Xu, Hongbin and Chen, Weitao and Zhou, Zhipeng and Xiao, Haihong and Sun, Baigui and Xie, Xuansong and Kang, Wenxiong},
  booktitle={Proceedings of the IEEE/CVF International Conference on Computer Vision (ICCV)},
  pages={14244--14253},
  year={2023},
  doi={10.1109/ICCV51070.2023.01314}
}

@inproceedings{wang2024embodiedscan,
  title={EmbodiedScan: A Holistic Multi-Modal 3D Perception Suite Towards Embodied AI},
  author={Wang, Tai and Mao, Xiaohan and Zhu, Chenming and Xu, Runsen and Lyu, Ruiyuan and Li, Peisen and Chen, Xiao and Zhang, Wenwei and Chen, Kai and Xue, Tianfan and Liu, Xihui and Lu, Cewu and Lin, Dahua and Pang, Jiangmiao},
  booktitle={Proceedings of the IEEE/CVF Conference on Computer Vision and Pattern Recognition (CVPR)},
  year={2024},
  doi={10.1109/CVPR52733.2024.01868},
}

@inproceedings{lin2025bip3d,
  title={BIP3D: Bridging 2D Images and 3D Perception for Embodied Intelligence},
  author={Xuewu Lin and Tianwei Lin and Lichao Huang and Hongyu Xie and Zhizhong Su},
  booktitle={Proceedings of the IEEE/CVF Conference on Computer Vision and Pattern Recognition (CVPR)},
  year={2025},
  pages={9007--9016},
  doi={10.1109/CVPR52734.2025.00842},
  eprint={2411.14869}
}

@article{xiao2024survey,
  title = {A Survey of Label-Efficient Deep Learning for 3D Point Clouds},
  author = {Xiao, Aoran and Zhang, Xiaoqin and Shao, Ling and Lu, Shijian},
  journal = {IEEE Transactions on Pattern Analysis and Machine Intelligence},
  year = {2024},
  doi = {10.1109/TPAMI.2024.3416302}
}

@article{jing2026,
title = {From geometric labels to semantic understanding of indoor building components using multimodal large language models},
journal = {Automation in Construction},
volume = {190},
pages = {107117},
year = {2026},
issn = {0926-5805},
doi = {https://doi.org/10.1016/j.autcon.2026.107117},
url = {https://www.sciencedirect.com/science/article/pii/S0926580526003584},
author = {Shuju Jing and Chao Yin},
}

\end{document}